\documentclass[10pt,twocolumn,letterpaper]{article}

\usepackage[algorithms]{wacv}

\usepackage{amsmath}
\providecommand{\linenumbers}{}
\providecommand{\nolinenumbers}{}
\usepackage{cuted}
\usepackage{booktabs}
\usepackage{placeins}
\usepackage{multirow}
\usepackage{array}
\usepackage{arydshln}
\usepackage{makecell}
\usepackage{pifont}
\usepackage{xcolor}
\usepackage{siunitx}
\usepackage{subcaption}
\usepackage{tikz}
\usetikzlibrary{spy,calc}
\usepackage{algorithm}
\usepackage{algpseudocode}
\AtEndPreamble{
    \crefname{figure}{figure}{figures}
    \Crefname{figure}{Figure}{Figures}
    \crefname{table}{table}{tables}
    \Crefname{table}{Table}{Tables}
}

\makeatletter
\define@key{Gin}{frame}[]{}
\makeatother

\newcommand{\comparisonrow}[2][0.187]{%
\includegraphics[width=#1\linewidth]{imgs/comparisons_seperate_images/#2/OriginalImage_outlined.jpg} &
\includegraphics[width=#1\linewidth]{imgs/comparisons_seperate_images/#2/MirrorFusion_outlined.jpg} &
\includegraphics[width=#1\linewidth]{imgs/comparisons_seperate_images/#2/FLUXFill_outlined.jpg} &
\includegraphics[width=#1\linewidth]{imgs/comparisons_seperate_images/#2/Qwen_outlined.jpg} &
\includegraphics[width=#1\linewidth]{imgs/comparisons_seperate_images/#2/Ours_outlined.jpg} \\
}

\newcommand{\mirrorbenchcomparisonrow}[2][0.187]{%
\includegraphics[width=#1\linewidth]{imgs/mirrorbench_v2_qualitative/#2/GT.jpg} &
\includegraphics[width=#1\linewidth]{imgs/mirrorbench_v2_qualitative/#2/MF2.0.jpg} &
\includegraphics[width=#1\linewidth]{imgs/mirrorbench_v2_qualitative/#2/FLUX.1Fill.jpg} &
\includegraphics[width=#1\linewidth]{imgs/mirrorbench_v2_qualitative/#2/Qwen.jpg} &
\includegraphics[width=#1\linewidth]{imgs/mirrorbench_v2_qualitative/#2/Ours.jpg} \\
}

\newcommand{\xmark}{\ding{55}}%

\newcommand{\MirrorFusion}{\text{MirrorFusion~2.0}}
\newcommand{\Qwen}{Qwen-Image-Edit}
\newcommand{\modelName}{FLUX.1 Fill}
\newcommand{\noiseInterpolationFunctionPower}{n}
\newcommand{\noisePredictionOriginalMask}{v'}
\newcommand{\noisePredictionSmallerMask}{v}
\newcommand{\firstTimestepToNoisePrediction}{t'}
\newcommand{\evaluationMethodName}{Reflection Consistency Score}
\newcommand{\firstTimestep}{T}
\newcommand{\timestep}{t}

\newcommand{\inputImageToFlux}{I}
\newcommand{\inputMaskToFlux}{M}

\newcommand{\fromsupp}[1]{#1}
\newcommand{\new}[1]{#1}

\definecolor{burntorange}{rgb}{0.81,.33,0}
\definecolor{darkgreen}{rgb}{0.1,0.6,0.1}
\definecolor{otherblue}{rgb}{0.1,0.4,0.8}

\newcommand{\suppsecImpl}{\cref{sec:supp_impl}}
\newcommand{\suppsecPseudocode}{\cref{sec:supp_pseudocode}}
\newcommand{\suppsecDatasets}{\cref{sec:supp_datasets}}
\newcommand{\suppsecHyperparam}{\cref{sec:supp_hyperparam}}
\newcommand{\suppsecRCSSweep}{\cref{sec:supp_rcs_sweep}}
\newcommand{\suppsecSeedFullMirror}{\cref{sec:supp_seed_variance_full_mirror}}
\newcommand{\suppsecQualitative}{\cref{sec:supp_more_qualitative}}
\newcommand{\suppsecRobustness}{\cref{sec:supp_robustness}}
\newcommand{\suppsecJitter}{\cref{sec:supp_jitter_psnr}}

\definecolor{wacvblue}{rgb}{0.21,0.49,0.74}
\usepackage[pagebackref,breaklinks,colorlinks,allcolors=wacvblue]{hyperref}

\title{Fill My Mirror: Geometry-Constrained Mirror Inpainting}

\author{
Ofek Basson \quad Shimon Vainer \quad Yacov Hel-Or \quad Ohad Fried\\
Reichman University
}

\begin{document}
\maketitle

\begin{tikzpicture}[remember picture, overlay]
\node[anchor=south, inner sep=0pt] at ([yshift=0.5in]current page.south) {%
  \parbox{0.9\textwidth}{\footnotesize\noindent This work has been submitted to the IEEE for possible publication. Copyright may be transferred without notice, after which this version may no longer be accessible.}%
};
\end{tikzpicture}

\nolinenumbers
\begin{strip}
\centering
\setlength{\tabcolsep}{2pt}
\renewcommand{\arraystretch}{1}
\begin{tabular}{ccccc}
\textbf{GT} &
\textbf{MF2.0} &
\textbf{FLUX.1Fill} &
\textbf{Qwen} &
\textbf{Ours} \\[0.1cm]
\comparisonrow{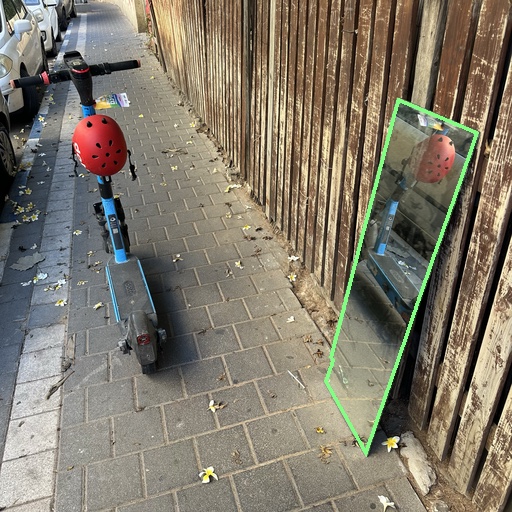}
\end{tabular}
\captionof{figure}{Mirror inpainting comparison. Generative baselines (\modelName, \Qwen) fill reflections that are geometrically inconsistent with the surrounding scene. \MirrorFusion~did not perform inpainting on this example, leaving the mirror region unchanged. Our geometry-constrained generative approach (right image) aligns reflected structures with the visible scene while preserving realism. The mirror region is outlined in green.}
\label{fig:teaser}
\end{strip}
\linenumbers

\begin{abstract}
Mirrors are common in real-world images, yet producing geometrically consistent reflections with generative models remains challenging. Unlike most objects, mirror appearance depends on scene geometry and viewpoint, making it hard to synthesize using learned appearance priors alone. We address this in the mirror inpainting setting, where the scene is fixed and only the mirror region is generated.

Our key insight is that much mirror content is geometrically constrained by the visible scene and need not be hallucinated. We estimate scene geometry and project visible content into the mirror to recover reflection regions determined by geometry. A generative model then completes the mirror region via a two-mask diffusion strategy balancing geometric constraints with the model's learned priors, reducing projection artifacts and improving reflection consistency. The method is training-free and applicable to complex real-world scenes.

We evaluate on MirrorBench-V2 (synthetic) and real images. Using standard and geometry-aware metrics, we show that explicitly using scene geometry improves consistency.
\end{abstract}

\section{Introduction}

Mirrors frequently appear in images generated by modern models, yet generating physically consistent reflections remains an unsolved challenge. Filling a mirror region in an image requires understanding the scene geometry and reasoning about parts of the scene that are not directly visible. We therefore focus on the task of mirror inpainting: given an image and a mirror mask, our goal is to generate reflection content that is geometrically consistent with the visible scene. Notably, this task differs from generating an image containing a mirror from scratch: when generating freely, a model can implicitly satisfy its learned priors by constructing a scene with simple or undemanding reflection geometry, whereas our inpainting setting fixes the scene and requires the reflection to be derived from actual geometry --- leaving no freedom to choose an easy arrangement.

\begin{figure}[t]
\centering
\setlength{\tabcolsep}{1pt}
\renewcommand{\arraystretch}{1.2}

\begin{tabular}{ccc}

\textbf{$0^\circ$} &
\textbf{$+5^\circ$} &
\textbf{$+10^\circ$} \\

\includegraphics[frame,
  width=0.31\linewidth,
  trim={0 125pt 0 125pt},
  clip
]{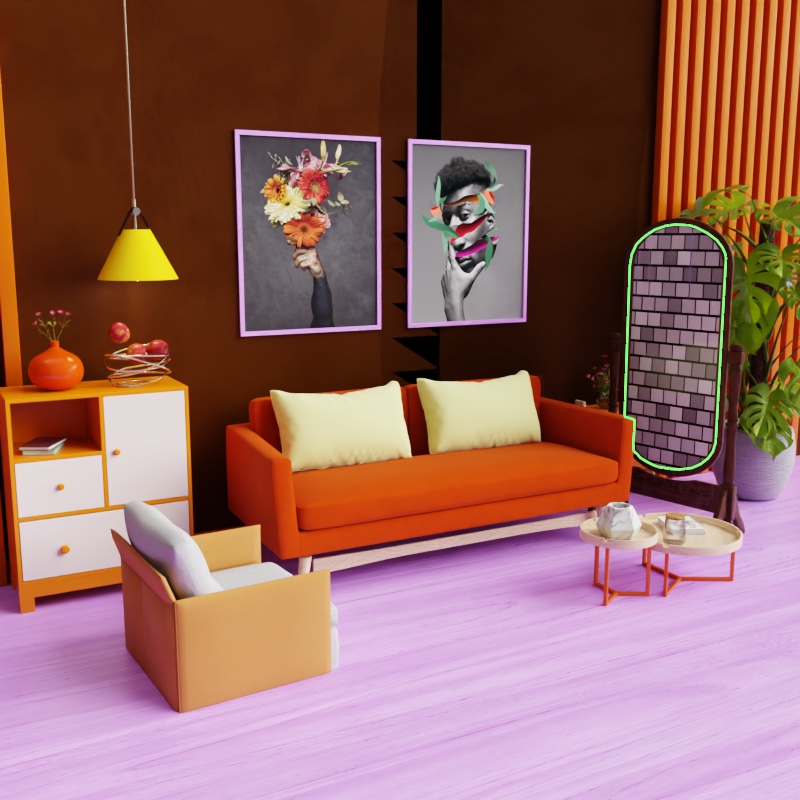} &
\includegraphics[frame,
  width=0.31\linewidth,
  trim={0 125pt 0 125pt},
  clip
]{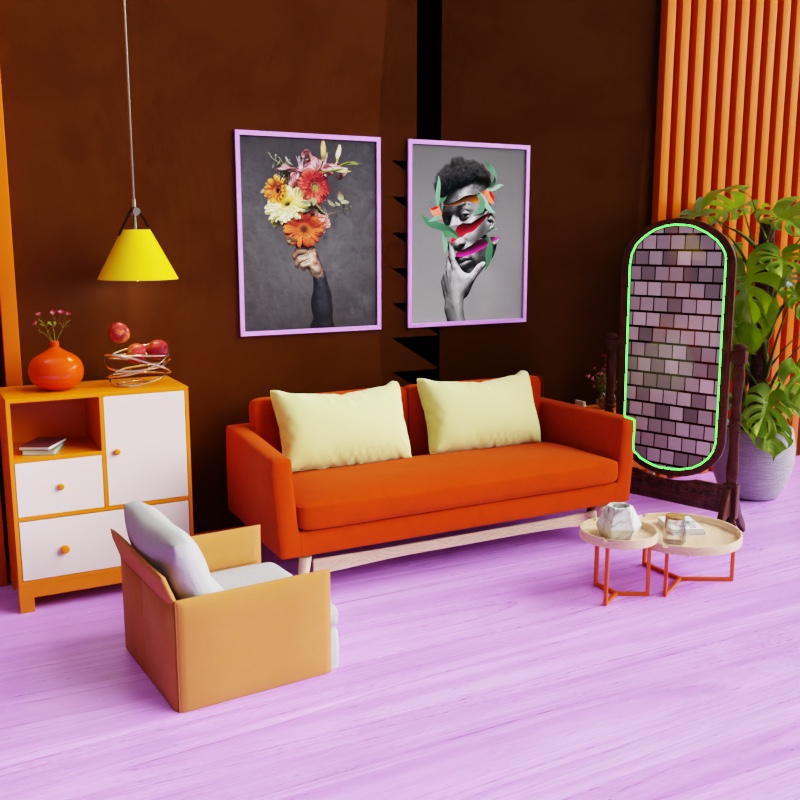} &
\includegraphics[frame,
  width=0.31\linewidth,
  trim={0 125pt 0 125pt},
  clip
]{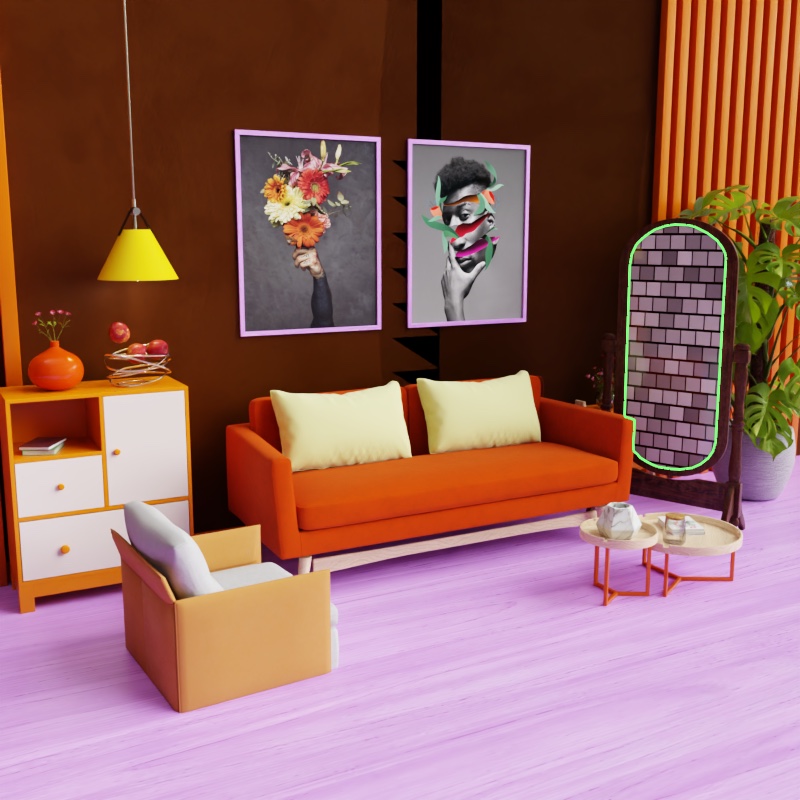} \\

\includegraphics[frame,
  width=0.31\linewidth,
  trim={0 125pt 0 125pt},
  clip
]{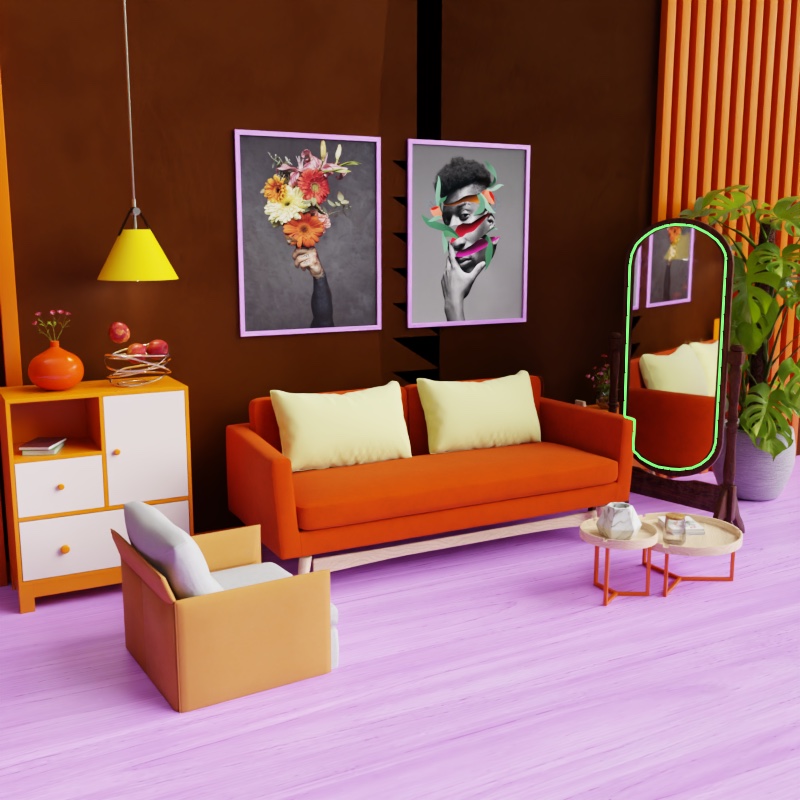} &
\includegraphics[frame,
  width=0.31\linewidth,
  trim={0 125pt 0 125pt},
  clip
]{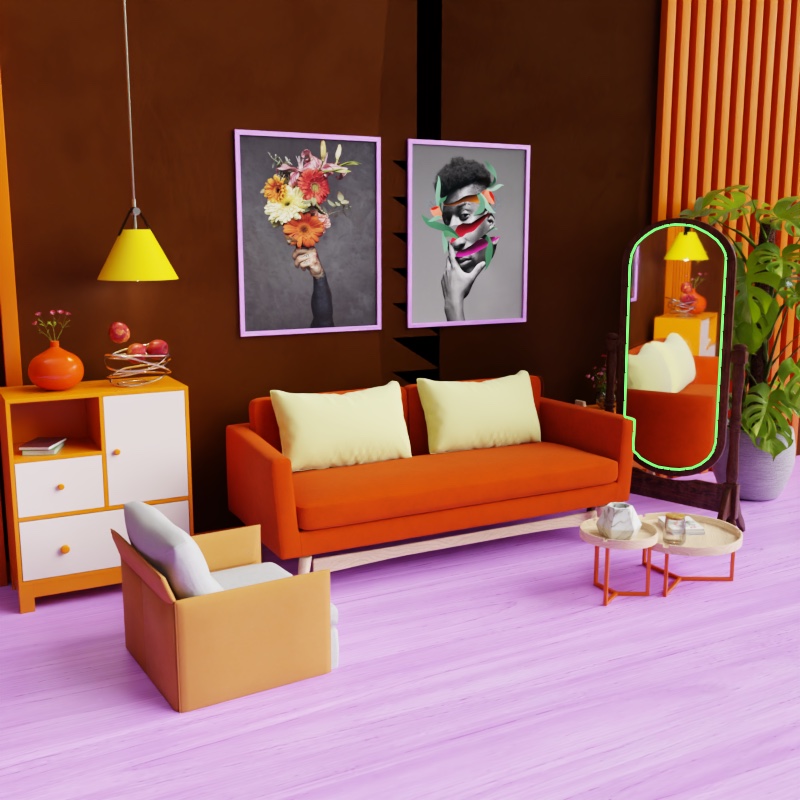} &
\includegraphics[frame,
  width=0.31\linewidth,
  trim={0 125pt 0 125pt},
  clip
]{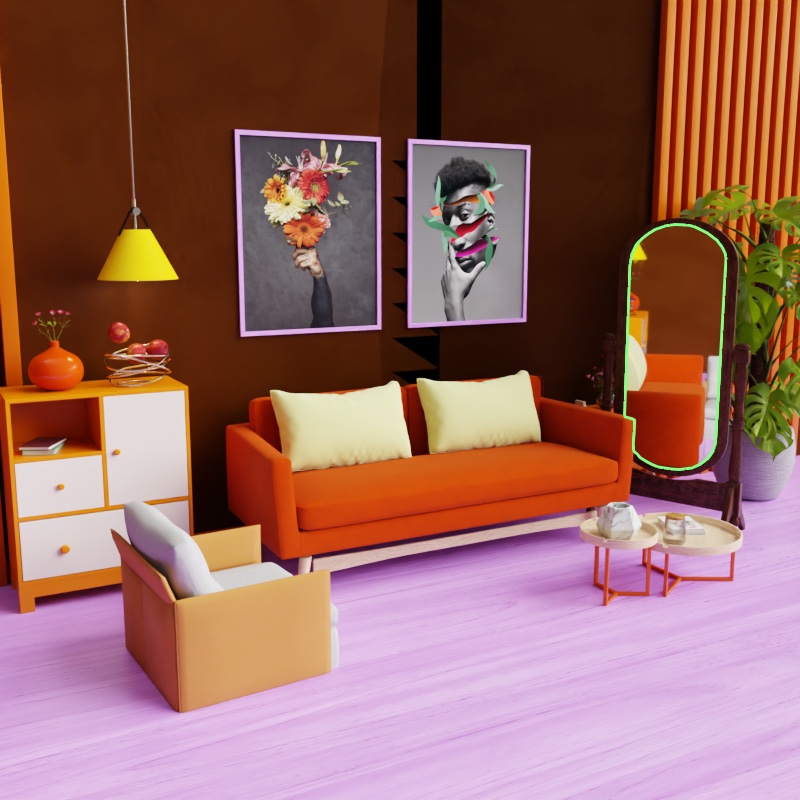} \\
\end{tabular}







\caption{
\textbf{Effect of small viewpoint changes on mirror appearance.} 
Columns show viewpoint changes for a non-reflective brick wall (top row) and a mirror (bottom row).
A $5^\circ$ change in orientation, which is nearly imperceptible to observers, causes a significant change in the mirror appearance, while the non-reflective textured surface changes only slightly. 
This highlights the strong geometric dependence of mirror appearance and emphasizes its fundamental difference from non-reflective objects.}
\label{fig:view_sensitivity}
\end{figure}

As illustrated in \cref{fig:teaser}, existing inpainting methods typically generate visually pleasing mirror content, but they fail to produce geometrically consistent reflections. One reason for this failure is that existing inpainting methods primarily rely on the learned priors of the model to hallucinate missing content in mirror regions. The fact that mirror appearance is not intrinsic to the surface, but is fully determined by the scene and viewing geometry, makes mirror behavior difficult to learn from image statistics alone: even small viewpoint changes can substantially alter the reflected content (\cref{fig:view_sensitivity}).

Recent works such as Reflecting Reality~\citep{reflecting} and MirrorVerse~\citep{dhiman2025mirrorverse} address this challenge by conditioning generation on depth information. While these approaches include some real images, they are predominantly trained on synthetic data, which affects their generalization to real-world scenes and diverse geometries. Moreover, depth is used only as an auxiliary input signal to guide hallucination, rather than as an explicit geometric anchor: to our knowledge, no prior mirror-generation method projects scene geometry into the mirror and uses that projection to directly constrain, rather than merely condition, the generated content.

Our key insight is that mirror reflections are largely determined by scene geometry and therefore do not need to be fully hallucinated. A significant portion of the mirror content can be recovered deterministically by projecting visible scene geometry into the mirror using estimated geometry. However, projection errors introduce artifacts that propagate into the final result, making purely projection-based completion insufficient. To address this, we adopt a two-mask diffusion formulation in which one mask governs inpainting of regions not constrained by geometry, while a second mask preserves and refines the projected reflection using the model's learned priors. This lets a single pre-trained diffusion model preserve, refine, or synthesize each region according to its geometric evidence, rather than hallucinating the whole mirror. The resulting method is \textbf{training-free} and improves with geometry quality.

We evaluate our method on MirrorBench-V2~\citep{dhiman2025mirrorverse}, a public benchmark providing ground-truth geometry, both with and without access to ground-truth geometry during inference, as well as on real images without ground-truth geometry. We compare our approach against strong baseline approaches using standard image-level metrics together with a newly introduced geometry-based reflection consistency evaluation. Our results show that explicitly incorporating scene geometry leads to better mirror reflections.

Our main contributions are:

\noindent\textbf{Geometry-constrained mirror inpainting.}\quad We introduce a training-free mirror inpainting method that explicitly decomposes the problem into (1) deterministic geometric scene projection and (2) generative completion of the remaining regions. To our knowledge, this is the first mirror-generation method to condition on explicit projected geometry rather than using depth only as an auxiliary signal. This formulation enforces physical reflection constraints while minimizing reliance on hallucination.

\noindent\textbf{Robust diffusion via two-mask decomposition.}\quad We propose a diffusion-time noise interpolation strategy that leverages two complementary masks to balance geometric constraints with the model's learned priors, mitigating artifacts caused by depth errors, occlusion ambiguities, and imperfect projections without retraining or modifying the underlying diffusion model.

\noindent\textbf{Geometry-aware evaluation without ground-truth geometry.}\quad We introduce a metric for evaluating the quality of mirror reflection in real images without ground-truth geometry, focusing mainly on mirror regions that are determined by the visible scene.

\section{Related Work}
Prior work related to our approach spans three directions: modeling mirror geometry in visual reconstruction and rendering, incorporating geometric constraints into generative image editing, and evaluating geometric consistency in image synthesis. We briefly review these lines of work and clarify how our method differs in explicitly enforcing reflection geometry during inference.

\subsection{Mirrors as a Special Case in Visual Modeling}

Mirrors violate the common assumption that an observed image corresponds directly to visible scene geometry, as they encode a virtual viewpoint rather than directly observed structure. As a result, mirror regions often introduce inconsistencies in reconstruction and rendering systems. This virtual-camera model of reflection has a long history in classical catadioptric stereo and self-calibration~\citep{gluckman2001catadioptric,sturm2006pose,rodrigues2010camera,ying2013self}, and underlies the neural approaches discussed below and our own projection stage.

Mirrors remain a persistent failure case in image and video synthesis~\citep{borji2025mirrors}, observed across multiple lines of work. In neural rendering and 3D reconstruction, mirrors often lead to duplicated geometry or inconsistent multi-view predictions. To address this, several methods~\citep{liu2024mirrorgaussian,Wang_2024_BMVC,van2025nerfs,wu2026reflect3r} explicitly model mirror geometry or introduce virtual reflected cameras, treating reflections as secondary viewpoints governed by mirror symmetry. Other works separate reflected and transmitted scene components via decomposition or reflection-supervised losses~\citep{zhang2024refgaussian,poirier2025editable,CHEN2025101277}.

While most of these approaches operate in multi-view reconstruction or neural rendering settings, they highlight a broader principle: mirror reflections require explicit geometric reasoning. Our work builds on this insight in the setting of single-image inpainting, where we enforce reflection geometry directly at inference time rather than relying solely on learned appearance priors.

\subsection{Geometry-Guided Image Editing and Inpainting}

A growing body of work incorporates geometric information into diffusion-based image editing and inpainting pipelines. Methods such as ControlNet~\citep{Zhang_2023_ICCV} and subsequent geometry-guided editing approaches inject spatial structure, including depth maps or surface normals, into the diffusion network to improve structural alignment. This line of work is enabled by rapid progress in monocular geometry estimation: Depth Anything~\citep{yang2024depthanything,yang2024depthanythingv2,depthanything3} and MoGe/MoGe-2~\citep{wang2025moge,wang2025moge2accuratemonoculargeometry} now provide standard building blocks for lifting single images into 3D, including for depth-conditioned editing and reflection synthesis.

More closely related to our setting, several inpainting approaches have been trained specifically to generate physically consistent reflections. \citet{reflecting} formulate reflection synthesis as a depth-conditioned inpainting task trained on a large-scale synthetic dataset. A subsequent extension~\citep{dhiman2025mirrorverse} improves generalization through targeted data augmentation and curriculum training strategies; both methods obtain depth from a monocular estimator and feed it to the network as an additional conditioning input. In both cases, geometric information influences the predicted noise at each diffusion step, but reflection consistency is learned implicitly through training rather than explicitly enforced at inference time, leaving room for geometrically inconsistent reflections when the learned prior dominates.

In contrast, our method is training-free and does not require mirror-specific examples. Instead, we impose deterministic geometric projection before the denoising process and use diffusion mostly to complete regions that are not constrained by reflection geometry.

\subsection{Evaluating Geometric Consistency of Mirror Content}

Existing mirror inpainting works~\citep{reflecting, dhiman2025mirrorverse} are typically evaluated using image-level reconstruction and perceptual metrics such as PSNR, SSIM, and LPIPS over the mirror region, along with CLIP similarity computed on the full image. While these metrics capture appearance similarity, they do not explicitly assess whether the generated reflection is geometrically consistent with the scene. Evaluating over the entire mirror mask mixes regions whose appearance is largely determined by scene geometry with regions that admit multiple visually plausible completions, reducing sensitivity to reflection correctness. To partially address geometry, \citet{reflecting} introduce a segmentation-based IoU metric over reflected object silhouettes; however, this proxy captures only coarse shape and cannot guarantee physically correct reflections. For example, two reflections may share similar outlines while differing in pose, scale, or viewpoint, leading to physically inconsistent mirror appearance despite high IoU. The follow-up work~\citet{dhiman2025mirrorverse} removes this geometry proxy and relies primarily on perceptual similarity within the mirror region, additionally selecting the best sample among multiple generated candidates based on similarity over the scene region, which can bias evaluation. Relatedly, \citet{yin2025refracting} evaluate physical realism for \underline{refractive} objects using masked structural metrics restricted to pixels whose appearance is constrained by ray tracing; however, this evaluation requires ground-truth geometry and cannot be applied to real images.

In contrast, we propose reflection-specific geometric assessment that focuses only on mirror pixels whose appearance is determined by scene geometry, and that can be estimated even without ground-truth depth.

\section{Method}
\label{sec:method}

\begin{figure*}[t]
    \centering
    \includegraphics[
      width=0.92\textwidth,
      trim={1.9cm 5.4cm 1.9cm 5.4cm},
      clip
    ]{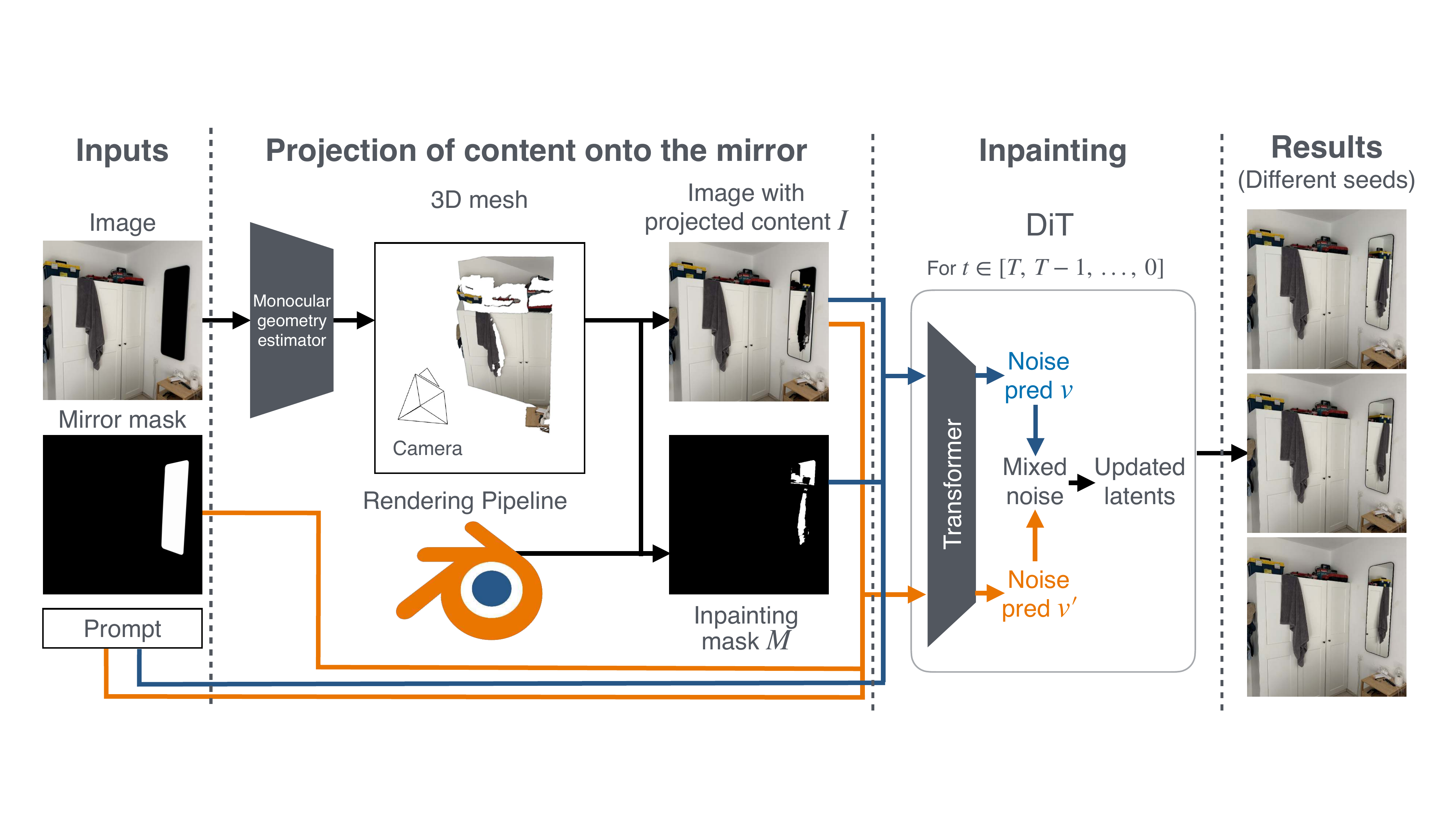}
    \caption{
    \textbf{Overview of the proposed method.} We first estimate a 3D mesh and camera parameters from the input image using a monocular geometry estimator.
    Using the reconstructed 3D geometry, we project the correct reflected content onto the mirror region, from which we derive the inpainting mask corresponding to the remaining regions to be filled. 
    During diffusion inpainting, two mask-guided noise predictions are computed and interpolated to balance geometric constraints with generative refinement (\cref{eq:noise_interpolation}); blue arrows denote the projection-derived geometry constraint mask; orange arrows denote the generative refinement mask corresponding to the full mirror region.
}
    \label{fig:overview}
\end{figure*}

Given a masked input image in which the mirror region is blank, a corresponding mirror mask, and a textual prompt guiding appearance generation, our method reconstructs mirror reflections using a geometry-aware inpainting pipeline (\cref{fig:overview}). We first leverage 3D scene attributes (ground-truth or estimated with a monocular geometry estimator~\citep{wang2025moge2accuratemonoculargeometry,depthanything3}) to project visible scene content into the mirror in a geometrically consistent manner (\cref{sec:projection}), identifying reflection pixels that can be deterministically recovered from scene geometry while leaving occluded or uncertain regions unresolved.

To fill the remaining regions, we introduce two complementary masks: a \emph{geometry constraint mask} derived from the projection and a \emph{generative refinement mask} corresponding to the full mirror region. During diffusion-based inpainting, interpolating noise predictions guided by these masks balances geometric consistency with the model's learned priors, reducing projection artifacts and improving reflection realism (\cref{sec:inpainting}); \suppsecPseudocode~gives the full inference pipeline.

\subsection{Projection of Content From Scene Onto the Mirror}
\label{sec:projection}
We first recover mirror content geometrically, constraining generation and reducing the inpainting mask.

Given an input masked image, we estimate geometry as well as the camera parameters using a monocular geometry estimator~\citep{wang2025moge2accuratemonoculargeometry,depthanything3} and reconstruct a 3D mesh. Using the mirror mask, we determine the mirror plane by fitting a 3D plane to the set of 3D points originating from the masked region, orienting the mirror normal towards the camera. On MirrorBench only, the geometry estimator occasionally returned non-finite depth values for mirror pixels; when fewer than $1\%$ of mirror points have a finite 3D value, we skip the projection and fall back to direct inpainting of the full mirror region. We reflect the reconstructed 3D scene across the estimated mirror plane and render the reflected geometry using Blender~\citep{blender}, supporting arbitrary mirror orientations; only scene content between the camera origin and the mirror plane is considered, and back-facing polygons are rendered black to model occlusions.

The resulting rendering produces both the reflected view and a mask indicating regions that remain unobserved. We remove content outside the mirror using the mirror mask, yielding the final inpainting mask $\inputMaskToFlux$ (the \emph{geometry constraint mask}) and the input image $\inputImageToFlux$, obtained by compositing the rendered reflection into the masked mirror region.

\subsection{Geometry-Constrained Inpainting via Two-Mask Decomposition}
\label{sec:inpainting}
After projecting known scene content onto the mirror, we fill the remaining regions. We instantiate our approach using \modelName~\citep{flux1fill}, a publicly available masked inpainting model. Applying this pipeline directly introduces artifacts from the geometric projection stage: (1) \emph{occlusion recovery failures}, where scene regions that should occlude others in the reflection are not visible in the input and therefore not reconstructed; and (2) \emph{inaccurate geometry estimates}, whose projection errors propagate because \modelName~fills only unprojected regions (\cref{fig:noise_interpolation_effect}).

To address projection artifacts while preserving geometric constraints, we model mirror synthesis using two complementary masks. The projected inpainting mask defines a \emph{geometry constraint mask}, covering only mirror pixels whose reflection could not be recovered from scene geometry. The original mirror mask defines a \emph{generative refinement mask} spanning the entire mirror region, representing the model's unconstrained generative prior and allowing the diffusion process to correct projection errors and complete missing or uncertain reflection content.

We implement this decomposition during diffusion by computing two velocity predictions for the rectified-flow denoiser at each denoising step $\timestep$, differing only in the inpainting mask provided to the network: $\noisePredictionSmallerMask$, obtained using the geometry constraint mask $\inputMaskToFlux$, and $\noisePredictionOriginalMask$, obtained using the generative refinement mask. The interpolated prediction is
\begin{equation}
\operatorname{mixed\ noise} =
\begin{cases}
\noisePredictionSmallerMask, & \timestep > \firstTimestepToNoisePrediction \\
\gamma^{\noiseInterpolationFunctionPower}\noisePredictionSmallerMask + \left(1-\gamma^{\noiseInterpolationFunctionPower}\right)\noisePredictionOriginalMask, & \timestep \le \firstTimestepToNoisePrediction
\end{cases}
\label{eq:noise_interpolation}
\end{equation}
where \firstTimestep~is the largest timestep and $\gamma = \timestep / \firstTimestep$.

This formulation reflects the complementary roles of the two masks across diffusion timesteps. Early steps determine global reflection structure, where the geometry constraint mask provides strong physical guidance, while later steps refine appearance details and benefit from the generative refinement mask and the model priors. Interpolating between the two predictions thus enforces structural consistency while allowing flexible completion of unconstrained regions. We normalize the mixed prediction to match the $\ell_2$ norm of $\noisePredictionSmallerMask$, preserving the correct noise scale, and restrict mixing to timesteps $\timestep \le \firstTimestepToNoisePrediction$ since interpolating across all timesteps empirically degraded results. The exponent $\noiseInterpolationFunctionPower$ controls transition sharpness: raising $\noiseInterpolationFunctionPower$ keeps the geometry-constraint mask's weight $\gamma^{\noiseInterpolationFunctionPower}$ close to zero for most of the remaining denoising steps, so the mixed prediction is generative-refinement--dominated for longer, with the weight rising sharply only in the first few steps after mixing begins at $\firstTimestepToNoisePrediction$ --- a sharper switch from the geometry-constraint to the generative-refinement mask than a lower exponent would give. This gives the model more freedom later but can weaken geometric adherence.

To quantitatively select the interpolation parameters $\noiseInterpolationFunctionPower$ and $\firstTimestepToNoisePrediction$, we use a two-stage criterion. First, we form a safe pool of configurations whose constrained-pixel masked PSNR against the GT image is within 0.5\,dB of the best configuration. This filters out settings that provide too much generative freedom and therefore deviate from the geometric constraints. Second, within this safe pool, we select the configuration with the lowest constrained-pixel masked PSNR against the initial projected image, which indicates the largest departure from the projection while remaining close to the GT-constrained solution. The full ablation and plots are provided in \suppsecHyperparam. This yields $\noiseInterpolationFunctionPower{=}13$ and $\firstTimestepToNoisePrediction{=}625$ as our final configuration, which we use in all reported experiments.

Our method assumes the backbone inpainting model accepts an explicit mask as part of the network input to the transformer. While newer models such as FLUX.2 and \Qwen\ are available, their inpainting inference follows a latent-compositing approach: the transformer operates without explicit mask conditioning, and the inpainted region is blended into the latent at each step after prediction. Our dual-mask approach is therefore incompatible because it requires two predictions through one mask-conditioned network. These models may also leave masks unchanged.

\section{Experiments}
\label{sec:experiments}

\paragraph{Datasets.} For synthetic data, we use MirrorBench-V2, the benchmark introduced alongside MirrorVerse~\citep{dhiman2025mirrorverse}, which provides ground-truth mirror masks and geometry. We run inference both with and without access to ground-truth geometry, while using the ground-truth geometry for evaluation in both settings. Since MirrorBench-V2 does not reflect real-world complexity, we complement it with 50 real captured images with manually annotated masks and prompts, covering diverse environments and mirror placements. Existing real-world mirror datasets (MSD~\citep{yang2019mirror}, DLSU-OMRS~\citep{gonzales2023dlsu}, PMD~\citep{lin2020pmd}) target segmentation and are unsuitable for quantitative inpainting evaluation (see \suppsecDatasets); we additionally use a small validation set of 15 Blender-rendered images to tune hyperparameters (see \suppsecHyperparam); data and prompts are available at \url{https://github.com/OfekBasson/Fill-My-Mirror}, which also links to the datasets on Hugging Face.

\paragraph{Baselines.} We compare against three representative paradigms: (1) Mask-based inpainting without geometric intervention tests learned priors alone and corresponds to removing our projection stage; we use \modelName~\citep{flux1fill}. (2) General-purpose image editing tests mirror completion without explicit geometry; we use \Qwen~\citep{wu2025qwenimagetechnicalreport}. (3) Depth-conditioned, geometry-aware mirror inpainting represents prior geometric approaches; we use \MirrorFusion~\citep{dhiman2025mirrorverse}.

\paragraph{Metrics.} Mirror pixels are either constrained by visible scene content, with a well-defined reference, or unconstrained because they reflect unseen content and admit multiple plausible completions. We report PSNR, SSIM, and LPIPS over the full mirror and over constrained pixels, obtained from ground truth on synthetic scenes or estimated via \evaluationMethodName~otherwise (\cref{sec:reflection_consistency_score}); we also report full-image CLIP similarity. On MirrorBench-V2, only $15\%$ of mirror pixels are constrained on average ($85\%$ unconstrained), which makes the constrained-pixel comparison the more meaningful evaluation.

\paragraph{Implementation Details.} See \suppsecImpl{}.

\subsection{Qualitative Results}

\begin{figure*}[t]
\centering
\setlength{\tabcolsep}{2pt}
\renewcommand{\arraystretch}{1}
\begin{tabular}{ccccc}
\textbf{GT} &
\textbf{MF2.0} &
\textbf{FLUX.1Fill} &
\textbf{Qwen} &
\textbf{Ours} \\[0.1cm]
\comparisonrow{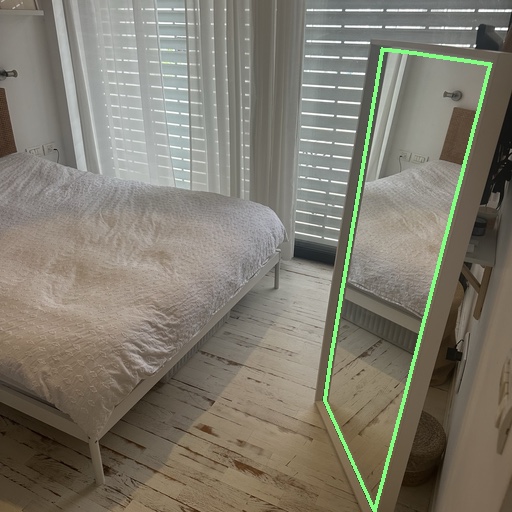}
\comparisonrow{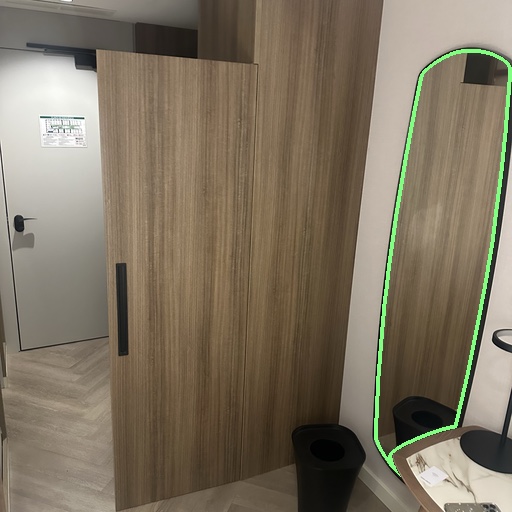}
\comparisonrow{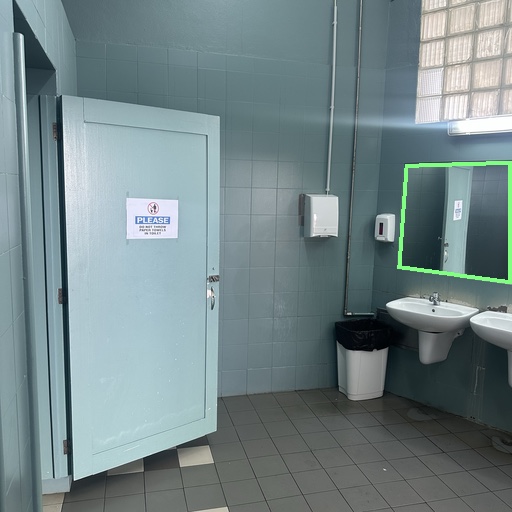}
\end{tabular}
\caption{
\textbf{Qualitative comparison of mirror inpainting results.} Columns (left to right): Ground Truth, \MirrorFusion, \modelName, \Qwen, and Ours.
In the first row, our method correctly reproduces the structure and position of the bed and headboard, closely matching the ground truth, whereas other methods exhibit structural distortions or misalignment.
In the second row, only our method accurately reflects the closet geometry and placement.
In the third row, our result correctly reflects the door in both color and position, while competing methods produce incorrect color, misalignment, or both. Additional results are provided in \suppsecQualitative.
}
\label{fig:qualitative_results}
\end{figure*}

\Cref{fig:qualitative_results} shows qualitative comparisons on real captured images without access to ground-truth geometry. Compared to baseline methods, our approach produces mirror reflections that better align with the visible scene geometry, preserving structural consistency across the mirror boundary, while competing methods often generate visually plausible reflections that exhibit misaligned structures or incorrect orientation. Additional qualitative results on both real images and MirrorBench-V2 are provided in \suppsecQualitative. These observations are consistent with the quantitative results below.

\subsection{Quantitative Results}
\label{sec:quantitative_results}

\subsubsection{Synthetic Scenes with Ground-Truth Geometry}
\Cref{tab:results_blender_images} compares methods on MirrorBench-V2~\citep{dhiman2025mirrorverse} with and without access to ground-truth geometry during inference, evaluated over all mirror pixels and over geometrically constrained pixels.
\begin{table}[t]
\centering
\caption{
\textbf{Quantitative comparison on synthetic scenes (MirrorBench-V2).}
PSNR, SSIM, and LPIPS are reported over \underline{constrained mirror pixels}, whose content is geometrically determined by the scene and identified using ground-truth geometry, and over the \underline{full mirror}.
}

\label{tab:results_blender_images}
\resizebox{\columnwidth}{!}{
\begin{tabular}{l c SSSSSS}
\toprule
\multirow{2.5}{*}{Method} & \multirow{2}{*}{\makecell{Geom. \\ Source}} & \multicolumn{3}{c}{Constrained Mirror Pixels}
& \multicolumn{3}{c}{Full Mirror Metrics} \\

\cmidrule(l{0.2em}r{0.2em}){3-5}
\cmidrule(l{0.2em}r{0.2em}){6-8}
& & {PSNR~$\uparrow$}
& {SSIM~$\uparrow$}
& {LPIPS~$\downarrow$}
& {PSNR~$\uparrow$}
& {SSIM~$\uparrow$}
& {LPIPS~$\downarrow$}
\\

\cmidrule(l{0.2em}r{0.2em}){1-1}
\cmidrule(l{0.2em}r{0.2em}){2-2}
\cmidrule(l{0.2em}r{0.2em}){3-3}
\cmidrule(l{0.2em}r{0.2em}){4-4}
\cmidrule(l{0.2em}r{0.2em}){5-5}
\cmidrule(l{0.2em}r{0.2em}){6-6}
\cmidrule(l{0.2em}r{0.2em}){7-7}
\cmidrule(l{0.2em}r{0.2em}){8-8}

\Qwen
& \xmark
& 10.5854
& 0.2417
& 0.1203
& 9.2732
& 0.2534
& 0.3438\\
\MirrorFusion
& GT
& 10.7515
& 0.2392
& 0.0585
& 9.8880
& 0.2539
& 0.2752\\
\modelName
& \xmark
& 14.8730
& 0.3153
& 0.0374
& 13.2097
& 0.3724
& \bfseries 0.2236\\
\hdashline
Ours
& Est.
& \bfseries 16.35
& \bfseries 0.37
& \bfseries 0.03
& \bfseries 13.81
& \bfseries 0.38
& \bfseries 0.2209\\
Ours
& GT
& \bfseries 21.92
& \bfseries 0.59
& \bfseries 0.02
& \bfseries 14.42
& \bfseries 0.44
& \bfseries 0.19\\
\bottomrule
\end{tabular}
}
\end{table}

Our method achieves better scores on all metrics. \modelName, which does not take depth as a conditioning input, still outperforms \MirrorFusion~due to its stronger prior. \Qwen~showed inconsistent behavior across experiments, in some cases leaving the mirror region unchanged, which can explain its poor quantitative results. When ground-truth geometry is provided at inference time (a fair comparison to \MirrorFusion), our results improve further. CLIP similarity remains identical (0.27) across methods, reflecting low sensitivity to localized mirror reconstruction differences.

\begingroup
\setcounter{figure}{5}
\begin{figure*}[t]
    \centering
    \includegraphics[
      width=0.95\textwidth
    ]{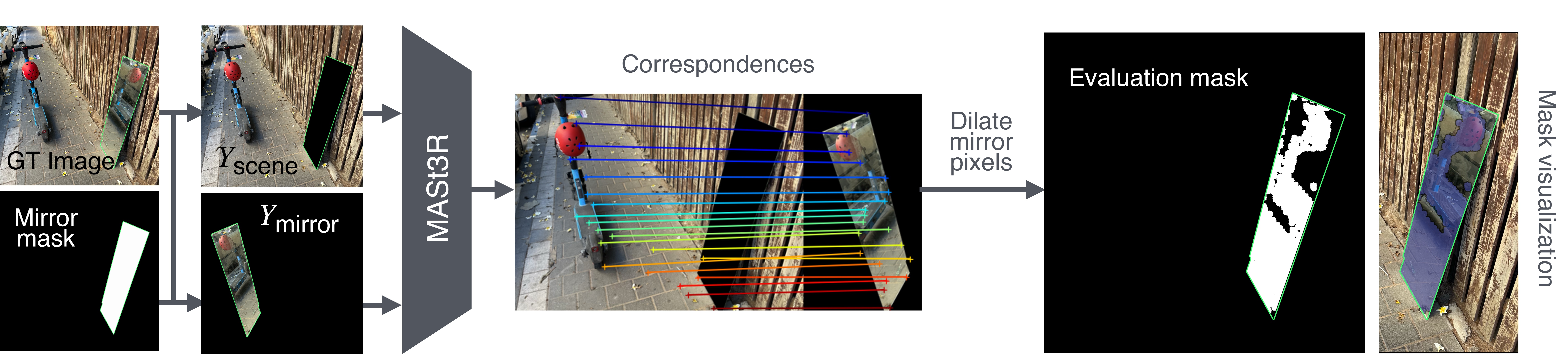}
    \caption{
    \textbf{\evaluationMethodName~(RCS) mask construction.} We decompose the input into a scene view and a horizontally flipped mirror view, estimate MASt3R correspondences with the two views constrained to share camera intrinsics, and dilate matched mirror pixels to form the evaluation mask. In this example, the scooter is visible in the input, so the corresponding portions of its reflection are geometrically constrained and included; regions reflecting unseen content admit multiple plausible completions and are excluded.
    }
    \label{fig:reflection_consistency_score}
\end{figure*}
\endgroup
\setcounter{figure}{4}

\subsubsection{Real Images without Ground-Truth Geometry}
\begin{table}[t]
\centering
\caption{
\textbf{Quantitative comparison on real images.}
PSNR, SSIM, and LPIPS are reported over \underline{constrained mirror pixels} identified by RCS (\cref{sec:reflection_consistency_score}) and over the \underline{full mirror}. All methods operate without ground-truth depth.
}
\label{tab:results_real_images}
\resizebox{%
    \ifdim\width>\columnwidth
        \columnwidth
    \else
        \width
    \fi}{!}{%
\begin{tabular}{l SSSSSS}
\toprule
\multirow{2.5}{*}{Method} & \multicolumn{3}{c}{\makecell{Constrained Mirror Pixels \\ (\evaluationMethodName)}}
& \multicolumn{3}{c}{Full Mirror Metrics} \\

\cmidrule(l{0.2em}r{0.2em}){2-4}
\cmidrule(l{0.2em}r{0.2em}){5-7}
& {PSNR~$\uparrow$}
& {SSIM~$\uparrow$}
& {LPIPS~$\downarrow$}
& {PSNR~$\uparrow$}
& {SSIM~$\uparrow$}
& {LPIPS~$\downarrow$}
\\

\cmidrule(l{0.2em}r{0.2em}){1-1}
\cmidrule(l{0.2em}r{0.2em}){2-2}
\cmidrule(l{0.2em}r{0.2em}){3-3}
\cmidrule(l{0.2em}r{0.2em}){4-4}
\cmidrule(l{0.2em}r{0.2em}){5-5}
\cmidrule(l{0.2em}r{0.2em}){6-6}
\cmidrule(l{0.2em}r{0.2em}){7-7}
\MirrorFusion
& 10.1178
& 0.3692
& 0.0701
& 10.0533
& 0.3903
& 0.1314
\\

\Qwen
& 10.1559
& 0.3422
& 0.0716
& 10.4413
& 0.3645
& 0.1284
\\

\modelName
& 13.7232
& 0.4548
& \bfseries 0.0513
& 13.4207
& 0.4723
& 0.1061
\\

\hdashline
Ours (no interpolation)
& 14.4849
& \bfseries 0.4810
& \bfseries 0.0478
& 13.5054
& \bfseries 0.4774
& \bfseries 0.1048
\\

Ours (with interpolation)
& \bfseries 14.6366
& \bfseries 0.4869
& \bfseries 0.0472
& \bfseries 13.62
& \bfseries 0.48
& \bfseries 0.10
\\

\bottomrule
\end{tabular}%
}
\end{table}

We compare our method against \modelName, \Qwen, and \MirrorFusion~on real images without ground-truth geometry (\cref{tab:results_real_images}), reporting PSNR, SSIM, and LPIPS over pixels estimated as geometrically constrained by \evaluationMethodName~(\cref{sec:reflection_consistency_score}) and over the full mirror region. Our method improves all metrics for both masks. \MirrorFusion~performs worst among all methods on real images --- a larger gap than on MirrorBench-V2 --- suggesting it overfits to its synthetic training distribution. The performance gap narrows on the full-mirror mask, as unconstrained regions admit multiple plausible hallucinations that add variance to pixel-based metrics, diluting the differences that RCS isolates by focusing on geometrically constrained pixels.

As a robustness check tolerating small local misalignment, \suppsecJitter{} compares standard constrained-region PSNR with jitter-tolerant PSNR; the results there further support our claims.

\begin{figure*}[t]

\centering
\begin{minipage}{0.04\linewidth}
\centering
\rotatebox{90}{}
\end{minipage}
\begin{minipage}{0.44\linewidth}
\centering
\textbf{Occlusion Recovery Failures}
\end{minipage}
\hfill
\begin{minipage}{0.44\linewidth}
\centering
\textbf{Geometry Estimation Errors}
\end{minipage}
\\[4pt] 
\begin{minipage}{0.04\linewidth}
\centering
\rotatebox{90}{\textbf{Projection}}
\end{minipage}
\begin{subfigure}[t]{0.22\linewidth}
\centering
\begin{tikzpicture} [baseline,
         spy using outlines={rectangle,
            magnification=2,
            width=1.672cm,
            height=1.936cm,
            connect spies
        }
    ]
        \node[inner sep=0pt] {
            \includegraphics[width=\linewidth]{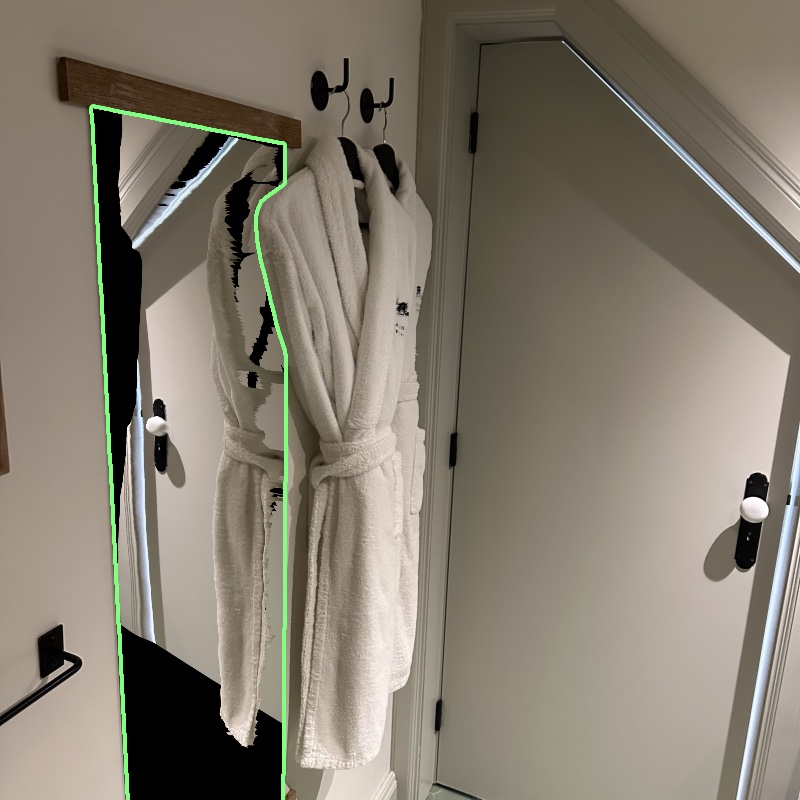}
        };

        \spy [blue] on (-0.66,-0.2) in node[below] at (1.07, 0.03);
\end{tikzpicture}
\end{subfigure}
\hfill
\begin{subfigure}[t]{0.22\linewidth}
\centering
\begin{tikzpicture} [baseline,
         spy using outlines={rectangle,
            magnification=2.3,
            width=1.672cm,
            height=1.936cm,
            connect spies
        }
    ]
        \node[inner sep=0pt] {
            \includegraphics[width=\linewidth]{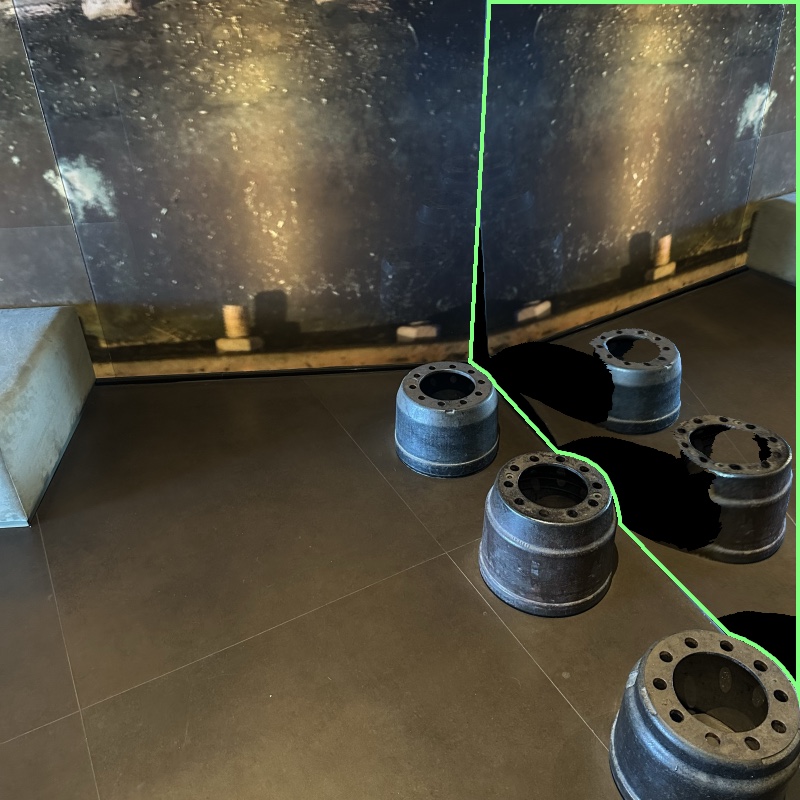}
        };

        \spy [blue] on (1.54,-0.4) in node[below] at (-1.07, 0.03);
\end{tikzpicture}
\end{subfigure}
\hfill
\begin{subfigure}[t]{0.22\linewidth}
\centering
\begin{tikzpicture} [baseline,
         spy using outlines={rectangle,
            magnification=3,
            width=1.672cm,
            height=1.936cm,
            connect spies
        }
    ]
        \node[inner sep=0pt] {
            \includegraphics[width=\linewidth]{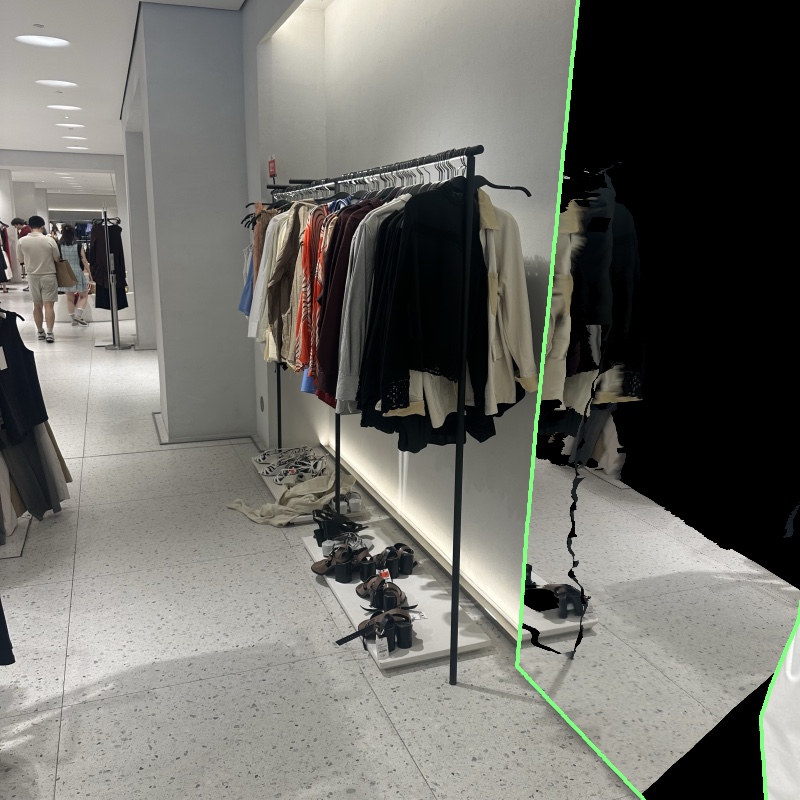}
        };

        \spy [blue] on (0.96,0.76) in node[below] at (-1.07, 0.03);
\end{tikzpicture}
\end{subfigure}
\hfill
\begin{subfigure}[t]{0.22\linewidth}
\centering
\begin{tikzpicture} [baseline,
         spy using outlines={rectangle,
            magnification=3,
            width=1.672cm,
            height=1.936cm,
            connect spies
        }
    ]
        \node[inner sep=0pt] {
            \includegraphics[width=\linewidth]{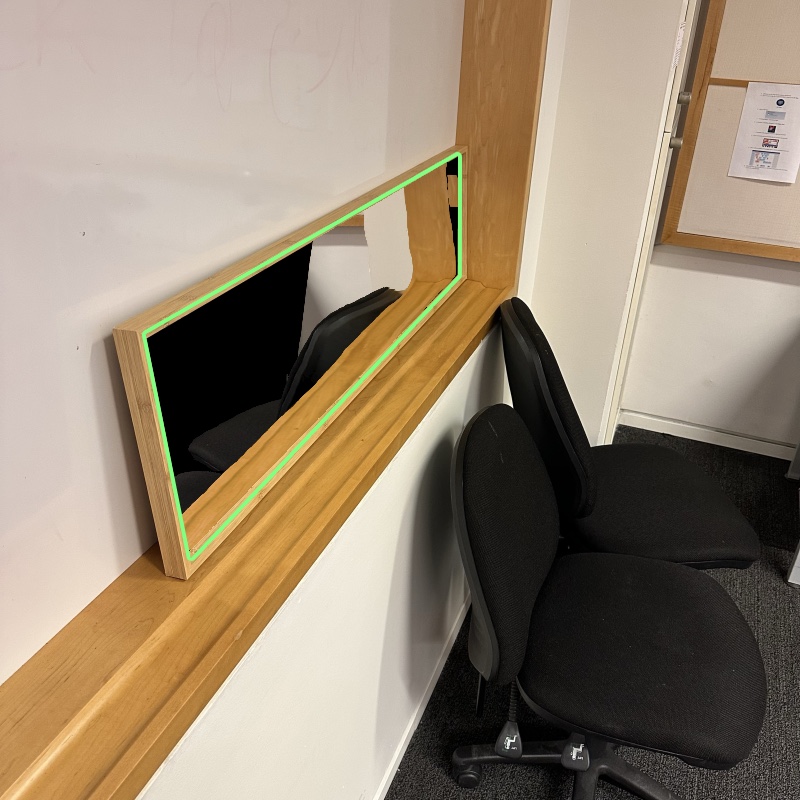}
        };

        \spy [blue] on (0.2,0.8) in node[below] at (-1.07, 0.03);
\end{tikzpicture}
\end{subfigure}
\\[4pt] 
\begin{minipage}{0.04\linewidth}
\centering
\rotatebox{90}{\textbf{No interp.}}
\end{minipage}
\begin{subfigure}[t]{0.22\linewidth}
\centering
\begin{tikzpicture} [baseline,
         spy using outlines={rectangle,
            magnification=2,
            width=1.672cm,
            height=1.936cm,
            connect spies
        }
    ]
        \node[inner sep=0pt] {
            \includegraphics[width=\linewidth]{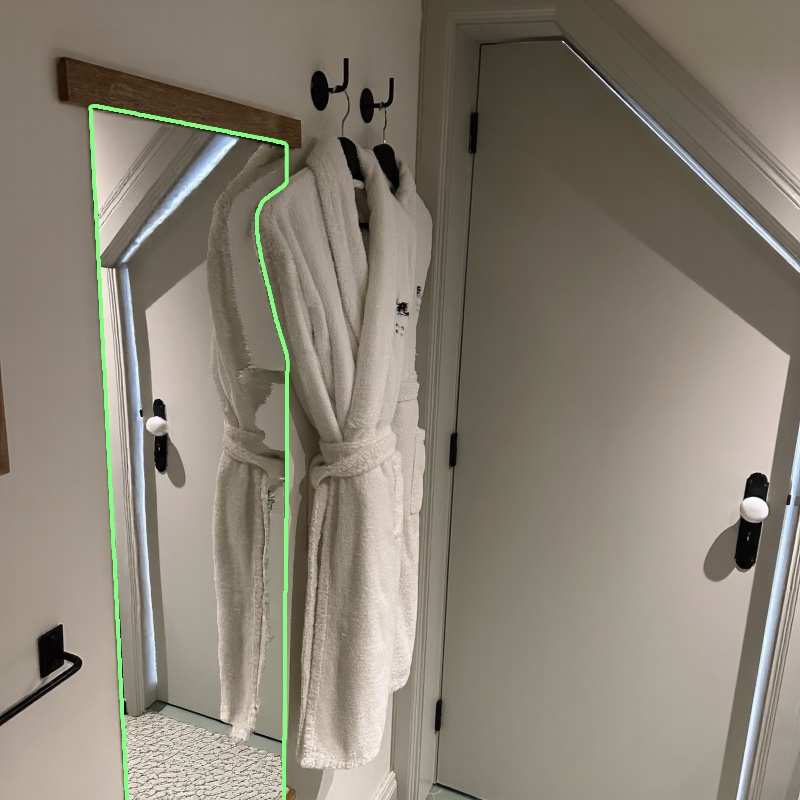}
        };

        \spy [blue] on (-0.66,-0.2) in node[below] at (1.07, 0.03);
\end{tikzpicture}
\end{subfigure}
\hfill
\begin{subfigure}[t]{0.22\linewidth}
\centering
\begin{tikzpicture} [baseline,
         spy using outlines={rectangle,
            magnification=2.3,
            width=1.672cm,
            height=1.936cm,
            connect spies
        }
    ]
        \node[inner sep=0pt] {
            \includegraphics[width=\linewidth]{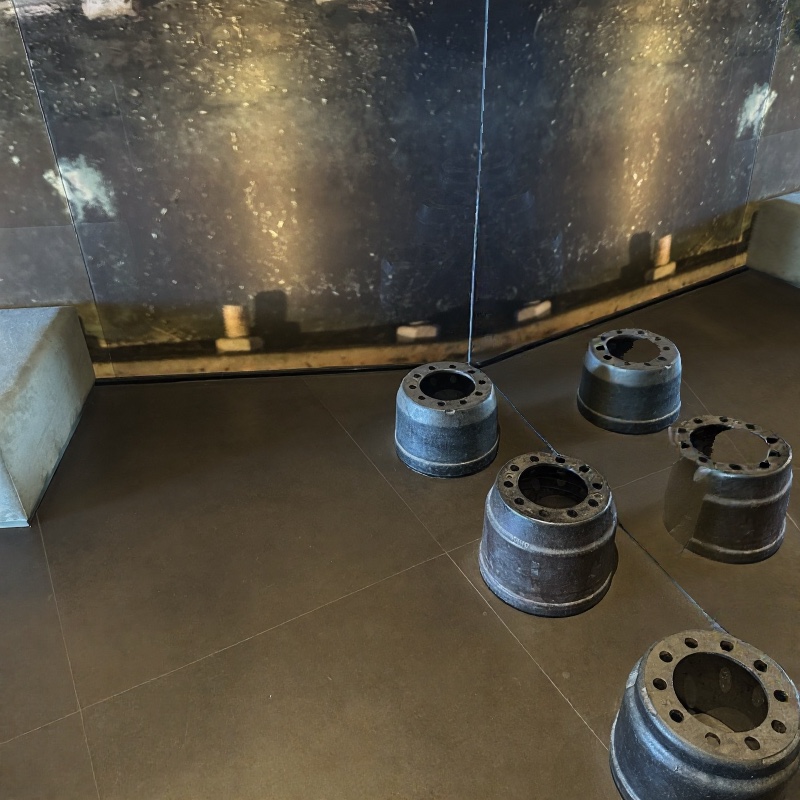}
        };

        \spy [blue] on (1.54,-0.4) in node[below] at (-1.07, 0.03);
\end{tikzpicture}
\end{subfigure}
\hfill
\begin{subfigure}[t]{0.22\linewidth}
\centering
\begin{tikzpicture} [baseline,
         spy using outlines={rectangle,
            magnification=3,
            width=1.672cm,
            height=1.936cm,
            connect spies
        }
    ]
        \node[inner sep=0pt] {
            \includegraphics[width=\linewidth]{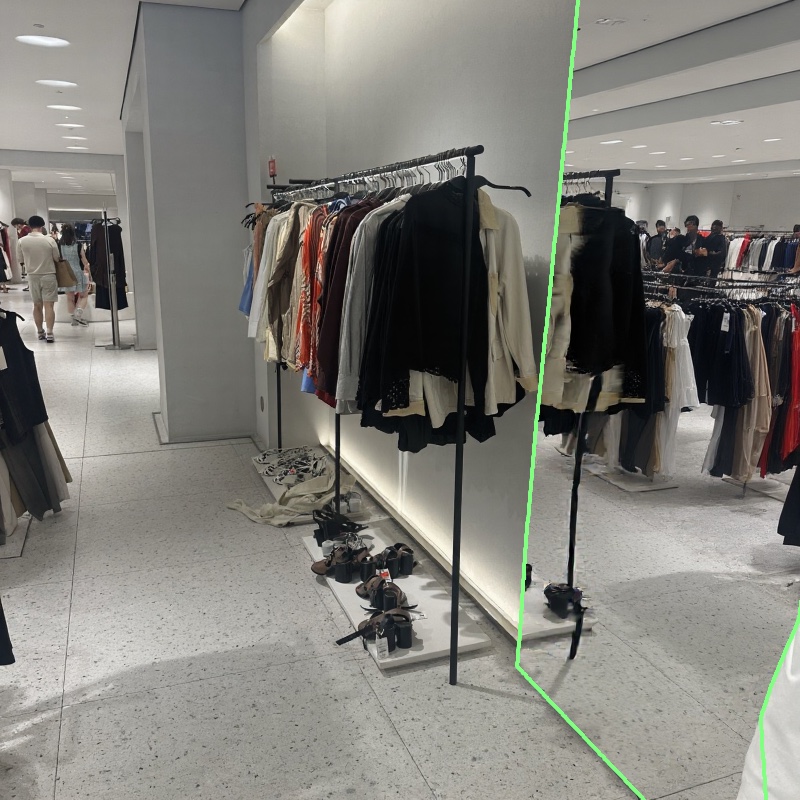}
        };

        \spy [blue] on (0.96,0.76) in node[below] at (-1.07, 0.03);
\end{tikzpicture}
\end{subfigure}
\hfill
\begin{subfigure}[t]{0.22\linewidth}
\centering
\begin{tikzpicture} [baseline,
         spy using outlines={rectangle,
            magnification=3,
            width=1.672cm,
            height=1.936cm,
            connect spies
        }
    ]
        \node[inner sep=0pt] {
            \includegraphics[width=\linewidth]{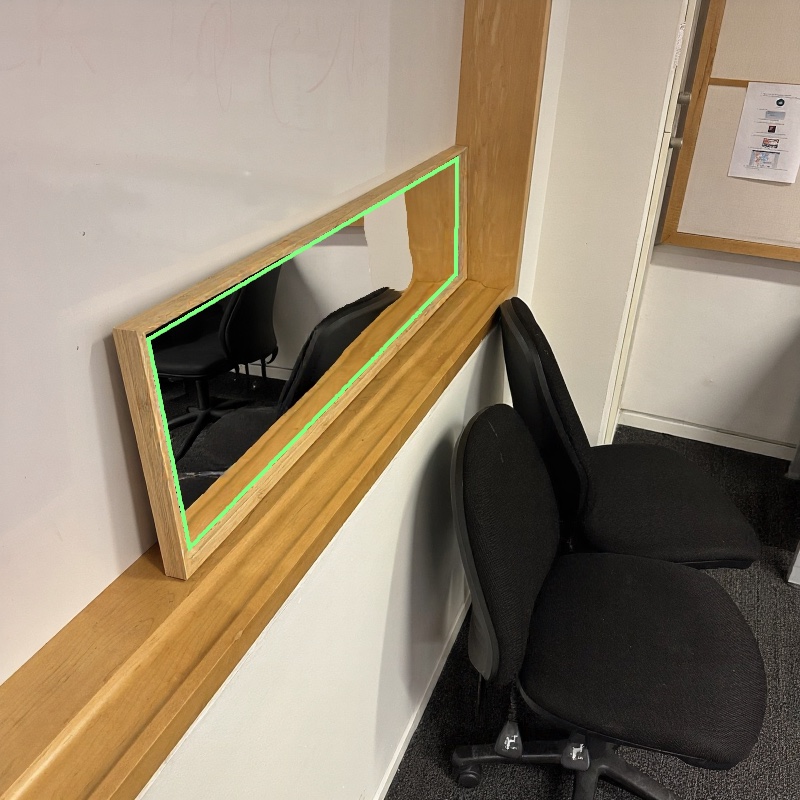}
        };

        \spy [blue] on (0.2,0.8) in node[below] at (-1.07, 0.03);
\end{tikzpicture}
\end{subfigure}
\\[4pt] 
\begin{minipage}{0.04\linewidth}
\centering
\rotatebox{90}{\textbf{With interp.}}
\end{minipage}
\begin{subfigure}[t]{0.22\linewidth}
\centering
\begin{tikzpicture} [baseline,
         spy using outlines={rectangle,
            magnification=2,
            width=1.672cm,
            height=1.936cm,
            connect spies
        }
    ]
        \node[inner sep=0pt] {
            \includegraphics[width=\linewidth]{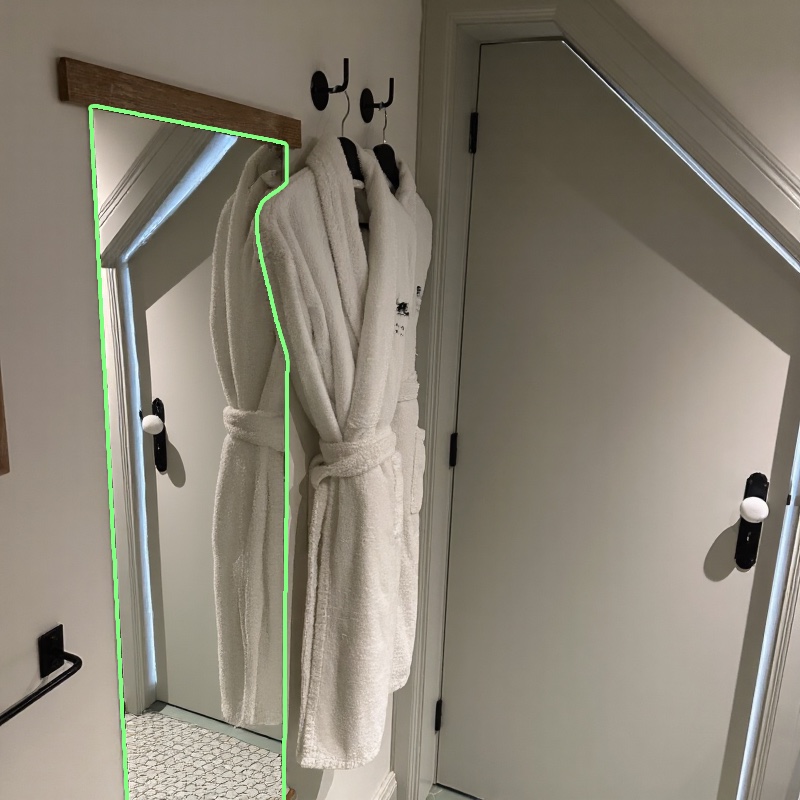}
        };

        \spy [blue] on (-0.66,-0.2) in node[below] at (1.07, 0.03);
\end{tikzpicture}
\end{subfigure}
\hfill
\begin{subfigure}[t]{0.22\linewidth}
\centering
\begin{tikzpicture} [baseline,
         spy using outlines={rectangle,
            magnification=2.3,
            width=1.672cm,
            height=1.936cm,
            connect spies
        }
    ]
        \node[inner sep=0pt] {
            \includegraphics[width=\linewidth]{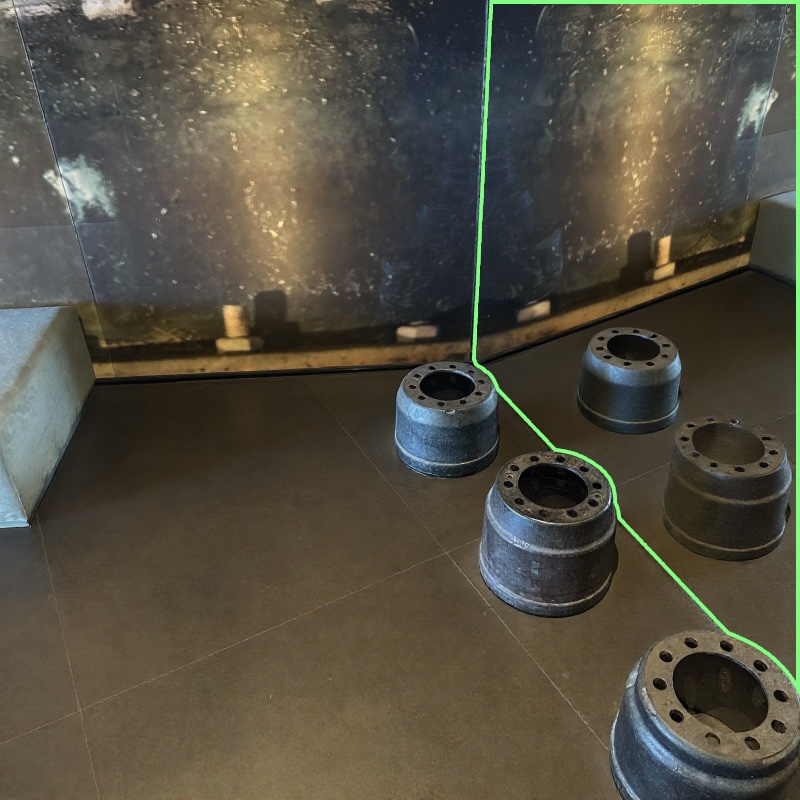}
        };

        \spy [blue] on (1.54,-0.4) in node[below] at (-1.07, 0.03);
\end{tikzpicture}
\end{subfigure}
\hfill
\begin{subfigure}[t]{0.22\linewidth}
\centering
\begin{tikzpicture} [baseline,
         spy using outlines={rectangle,
            magnification=3,
            width=1.672cm,
            height=1.936cm,
            connect spies
        }
    ]
        \node[inner sep=0pt] {
            \includegraphics[width=\linewidth]{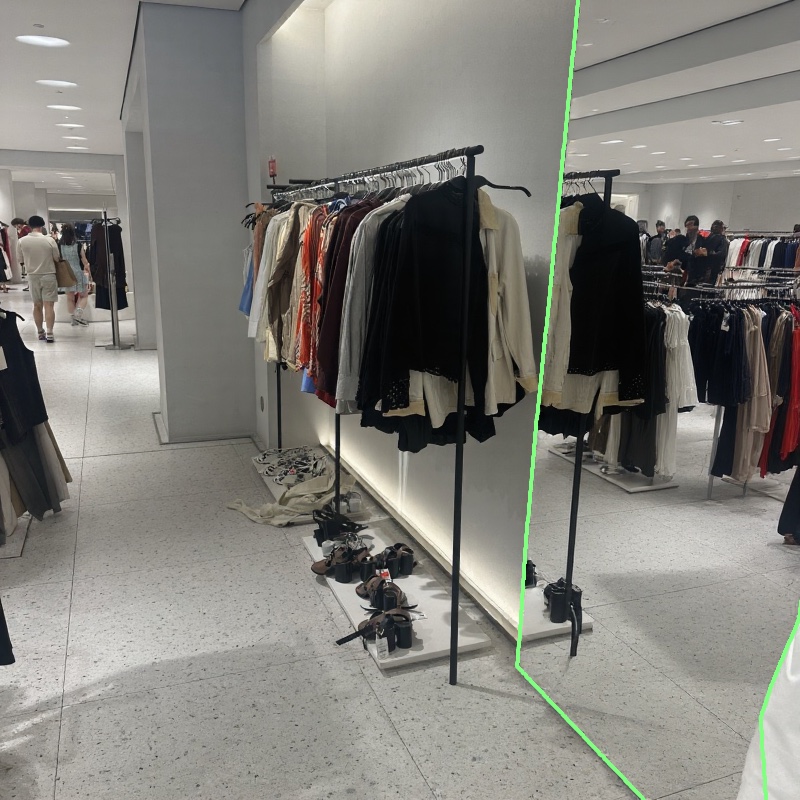}
        };

        \spy [blue] on (0.96,0.76) in node[below] at (-1.07, 0.03);
\end{tikzpicture}
\end{subfigure}
\hfill
\begin{subfigure}[t]{0.22\linewidth}
\centering
\begin{tikzpicture} [baseline,
         spy using outlines={rectangle,
            magnification=3,
            width=1.672cm,
            height=1.936cm,
            connect spies
        }
    ]
        \node[inner sep=0pt] {
            \includegraphics[width=\linewidth]{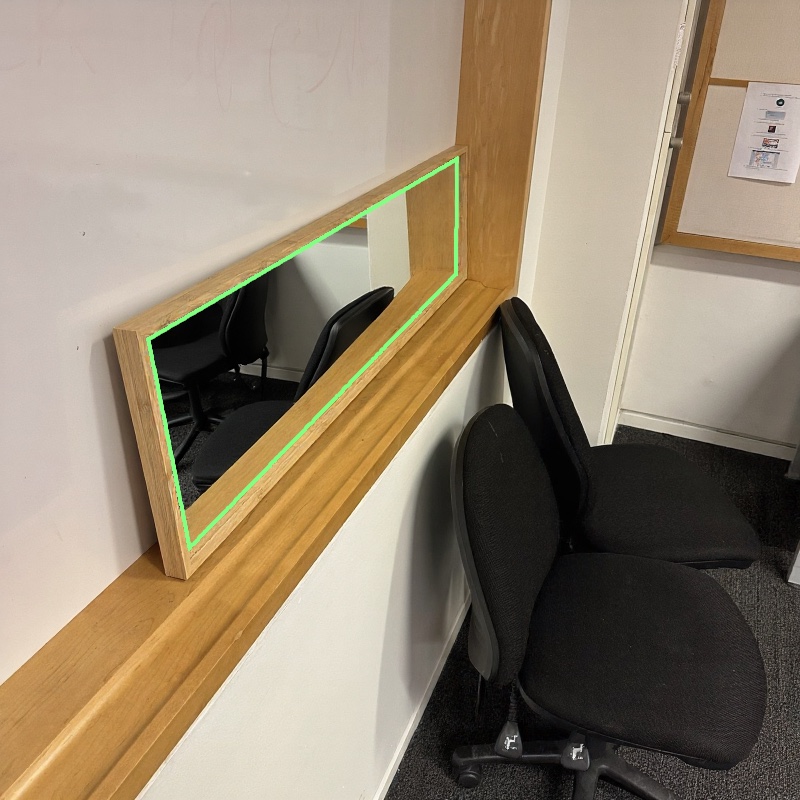}
        };

        \spy [blue] on (0.2,0.8) in node[below] at (-1.07, 0.03);
\end{tikzpicture}
\end{subfigure}
\caption{
\textbf{Failure cases and effect of noise interpolation.}
\underline{Columns:} Left: occlusion recovery failures due to missing geometry (door visible through bathrobe; floor visible inside cylinder). Right: geometry estimation errors causing structural distortions (smeared clothes reflection; uneven whiteboard frame with border gap).
\underline{Rows:} Top: projection only; Middle: projection + inpainting without noise interpolation (artifacts remain); Bottom: projection + inpainting with noise interpolation (ours), which resolves occlusion inconsistencies and reduces geometric distortions.}
\label{fig:noise_interpolation_effect}
\end{figure*}

We next ablate the dual-mask decomposition with noise interpolation: removing it degrades scores on almost all metrics (qualitative examples in \cref{fig:noise_interpolation_effect}). We verify these findings via paired significance tests (paired $t$-test and Wilcoxon signed-rank, Benjamini--Hochberg corrected across baselines per metric): the dual-mask decomposition yields statistically significant improvements over the no-interpolation ablation on all full-mirror and RCS-constrained metrics; improvements over \modelName~are significant on all RCS-constrained metrics; and improvements over \Qwen~and \MirrorFusion~are significant on all evaluated metrics.

Seed consistency results are provided in \suppsecSeedFullMirror. Our method not only outperforms baselines on both mask types but also produces substantially more stable outputs across seeds, suggesting our geometric conditioning anchors the output to physical scene constraints rather than relying on random generation.

\subsection{Robustness to Imperfect Geometry}
\label{sec:robustness_main}
To examine when noise interpolation helps most, we synthetically degrade the projection input depth and measure the resulting PSNR/SSIM gap between our method with and without interpolation. Given ground-truth depth $D_{\text{GT}}$ and MoGe-v2 estimate $D_{\text{est}}$, we degrade the depth as
\begin{equation}
    D_\lambda = (1-\lambda)\, D_{\text{GT}} + \lambda\, D_{\text{est}}, \qquad \lambda \in [0, 1.5],
    \label{eq:depth_degradation}
\end{equation}
and re-run our pipeline with $D_\lambda$ in place of the estimated geometry. As shown in \suppsecRobustness, the gap grows as geometry degrades (Pearson $r{=}0.741$ for PSNR, $r{=}0.778$ for SSIM): interpolation is mildly harmful with exact geometry but increasingly corrects occlusions and errors as geometry worsens.

\subsection{\evaluationMethodName}
\label{sec:reflection_consistency_score}

Without ground-truth geometry, we cannot directly isolate geometrically constrained mirror pixels for evaluation. We therefore propose \textbf{\evaluationMethodName} (RCS), illustrated in \cref{fig:reflection_consistency_score}. Following Reflect3r~\citep{wu2026reflect3r}, we decompose the input into a scene view and a horizontally flipped mirror view and use MASt3R~\citep{leroy2024grounding}, with the two views constrained to share camera intrinsics, to estimate dense pixel correspondences. Matched mirror pixels are dilated and intersected with the mirror region to form the evaluation mask, over which we compute masked PSNR, SSIM, and LPIPS. Parameter selection and full validation are provided in \suppsecRCSSweep.

Crucially, RCS is independent of our generation pipeline and does not reuse the signal that produced our results. Generation relies on monocular geometry estimation, mirror-plane fitting, and explicit Blender projection; RCS uses none of these components or outputs, constructing its mask solely from independent MASt3R correspondences. It therefore cannot favor our method by evaluating against its own geometry. On MirrorBench-V2, RCS achieves $0.72$ precision ($\sigma{=}0.16$) and $0.88$ recall ($\sigma{=}0.20$).

\section{Limitations}

The main limitation of our approach is its dependence on the accuracy of estimated scene geometry. When recovered geometry is inaccurate, the projected reflection may become misaligned, and although noise interpolation mitigates moderate errors, it cannot fully resolve large structural inconsistencies. Importantly, our framework is not tied to a particular geometry estimator, so it benefits as more accurate estimation methods become available. As the projected region shrinks, our pipeline increasingly relies on the inpainting model's generation prior, converging to its behavior when no projection is available.

Since our projection copies colors directly from the observed view, view-dependent effects such as specular highlights may appear incorrect in the reflection, and glass-window reflections --- which combine reflection with transmission --- are outside our scope.

In addition, our formulation assumes a single planar mirror. Handling multiple mirrors or higher-order inter-reflection effects would require extending the geometric projection stage, which we leave for future work. The method also does not address curved mirrors.

Adding geometry estimation and projection to inpainting raises runtime to 100.5s per image, versus 47.4s for \modelName~alone ($2.1\times$), at the same GPU memory (24.5GB).

\section{Conclusion}

We introduced Fill My Mirror, a training-free method for geometry-aware mirror inpainting that decomposes the problem into deterministic geometric projection and generative completion of the remaining regions, using a two-mask diffusion strategy with noise interpolation to balance physical reflection constraints with the model's learned priors and resolve occluded or uncertain regions.

Experiments on synthetic and real datasets show consistent improvements over inpainting, image editing, and mirror-aware generation baselines, particularly on geometrically constrained mirror pixels, and robustness to imperfect geometry. We further propose Reflection Consistency Score, a correspondence-based protocol for evaluating reflection correctness on real images without ground-truth geometry.

{
    \small
    \bibliographystyle{ieeenat_fullname}
    \bibliography{main}
}

\clearpage
\appendix
\setcounter{figure}{0}
\setcounter{table}{0}
\setcounter{algorithm}{0}
\renewcommand{\thetable}{\thesection\arabic{table}}
\renewcommand{\thefigure}{\thesection\arabic{figure}}
\renewcommand{\thealgorithm}{\thesection\arabic{algorithm}}

\begin{center}
{\Large\bf Supplementary Material}
\end{center}
\vspace{6pt}

This supplementary material provides implementation details and pseudocode; a discussion of existing mirror datasets, their limitations for inpainting, and MirrorVerse examples; the full noise-interpolation hyperparameter selection and its qualitative effect; RCS parameter selection and validation; seed-consistency analyses over both constrained pixels and the full mirror; additional qualitative comparisons on real images and MirrorBench-V2; robustness analysis under imperfect geometry; and a jitter-tolerant PSNR robustness check.

\section{Implementation Details}
\label{sec:supp_impl}

All experiments are conducted at a resolution of $1024 \times 1024$ pixels using 30 diffusion inference steps. All outputs are resized to the ground-truth resolution before computing evaluation metrics. For each method, we followed the official inference example provided in the model's source code as closely as possible. For \Qwen, the official example script uses \texttt{Qwen/Qwen-Image-Edit}; all reported results use the newer \texttt{Qwen/Qwen-Image-Edit-2511} weights, which we found to produce better results. All inference scripts are available at \url{https://github.com/OfekBasson/Fill-My-Mirror}.
The MirrorBench-V2 images were sampled with seed $0$; real images were filled using seeds $0, 42, 512$.

We provide implementations using two monocular geometry estimators: MoGe-v2~\citep{wang2025moge2accuratemonoculargeometry} and Depth Anything~3~\citep{depthanything3}. Both can be used as drop-in replacements within our pipeline; all reported results use MoGe-v2. On an RTX 6000 Ada, the geometric projection stage takes an average of 38 seconds per image on the real-image dataset. Adding this stage on top of inpainting brings total runtime to 100.5s per image, versus 47.4s for \modelName~alone ($2.1\times$), at the same peak GPU memory (24.5GB).

\section{Pseudocode}
\label{sec:supp_pseudocode}

\Cref{alg:pipeline} gives the full inference pipeline.

\begin{algorithm}[t]
\caption{Full Inference Pipeline}
\label{alg:pipeline}
\begin{algorithmic}[1]
\Require Masked image $I$, mirror mask $M_{\text{mirror}}$, text prompt $p$, interpolation start timestep $\firstTimestepToNoisePrediction$, exponent $\noiseInterpolationFunctionPower$, projection skip threshold $\tau{=}0.01$
\Ensure Inpainted image $\hat{I}$
\State Estimate geometry $G$ and camera parameters $K$ from $I$ \citep{wang2025moge2accuratemonoculargeometry,depthanything3}
\State Reconstruct 3D mesh $\mathcal{M}$ from $G$ and $K$
\If{$|\text{finite mirror points}| / |\text{mirror points}| < \tau$} \Comment{unreliable geometry; skip projection}
  \State $M_{\text{geo}} \leftarrow M_{\text{mirror}}$;\quad $\inputImageToFlux \leftarrow I$
\Else
  \State Fit plane $\Pi$ to mirror 3D points; orient normal toward camera
  \State Filter scene to points lying between the camera and $\Pi$
  \State Reflect filtered scene across $\Pi$; render image
  \State Obtain rendered reflection $R$ and projected-pixel mask $M_{\text{proj}}$
  \State $M_{\text{geo}} \leftarrow (M_{\text{mirror}} \setminus M_{\text{proj}}) \cap M_{\text{mirror}}$ \Comment{unprojected mirror pixels}
  \State $\inputImageToFlux \leftarrow I$ with mirror region composited with $R$
\EndIf
\For{each denoising step $t = T, \ldots, 0$}
  \State $\noisePredictionSmallerMask \leftarrow \mathrm{VelocityPred}(\inputImageToFlux, M_{\text{geo}}, p, t)$ \Comment{geometry-constrained}
  \If{$t > \firstTimestepToNoisePrediction$}
    \State $v_{\text{mix}} \leftarrow \noisePredictionSmallerMask$
  \Else
    \State $\noisePredictionOriginalMask \leftarrow \mathrm{VelocityPred}(\inputImageToFlux, M_{\text{mirror}}, p, t)$ \Comment{generative refinement}
    \State $\gamma \leftarrow t / \firstTimestep$
    \State $v_{\text{mix}} \leftarrow \gamma^{\noiseInterpolationFunctionPower}\noisePredictionSmallerMask + (1 - \gamma^{\noiseInterpolationFunctionPower})\noisePredictionOriginalMask$
    \State $v_{\text{mix}} \leftarrow v_{\text{mix}} \cdot \|\noisePredictionSmallerMask\|_2 / \|v_{\text{mix}}\|_2$ \Comment{preserve noise scale}
  \EndIf
  \State Update latent using $v_{\text{mix}}$
\EndFor
\State \Return decoded image $\hat{I}$
\end{algorithmic}
\end{algorithm}

\section{Existing Mirror Datasets and Their Limitations for Inpainting}
\label{sec:supp_datasets}

Publicly available real-world mirror datasets --- MSD~\citep{yang2019mirror}, DLSU-OMRS~\citep{gonzales2023dlsu}, and PMD~\citep{lin2020pmd} --- were designed for the mirror \emph{segmentation} task and are not directly suitable for quantitative inpainting evaluation. They share several limitations in this context:
\begin{itemize}
  \item \textbf{Ambiguous or invisible reflections.} Many images contain mirrors that do not visibly reflect any identifiable scene content, either due to camera angle, extreme foreshortening, or content that lies entirely outside the field of view. For such images, the reflection can be hallucinated in many equally valid ways, making pixel-level ground truth unreliable.
  \item \textbf{Non-planar and partial mirrors.} The datasets include curved mirrors and mirrors that are partially occluded or cropped, which violates the planar mirror assumption required for geometry-based evaluation.
  \item \textbf{No text captions.} None of the datasets include scene descriptions, which are required by all prompt-conditioned inpainting methods including ours.
\end{itemize}

To still make use of these established benchmarks, we evaluate our method on a subset of images from the MirrorVerse dataset~\citep{dhiman2025mirrorverse}, which is also based on MSD and provides generated GT reflections alongside text prompts. We use the prompts provided in that paper and compare against the MirrorVerse results and \modelName. As shown in \cref{fig:mirrorverse_examples}, our method produces more geometrically consistent reflections in the mirrors.

\newcommand{\mirrorverserow}[1]{%
\includegraphics[width=0.235\linewidth]{imgs/mirrorverse_examples/#1_gt.jpg} &
\includegraphics[width=0.235\linewidth]{imgs/mirrorverse_examples/#1_mirrorverse.jpg} &
\includegraphics[width=0.235\linewidth]{imgs/mirrorverse_examples/#1_flux.jpg} &
\includegraphics[width=0.235\linewidth]{imgs/mirrorverse_examples/#1_ours.jpg} \\
}

\begin{figure*}[!ht]
\centering
\setlength{\tabcolsep}{3pt}
\renewcommand{\arraystretch}{1}
\begin{tabular}{cccc}
\textbf{GT} &
\textbf{MirrorVerse} &
\textbf{FLUX.1 Fill} &
\textbf{Ours} \\[0.1cm]
\mirrorverserow{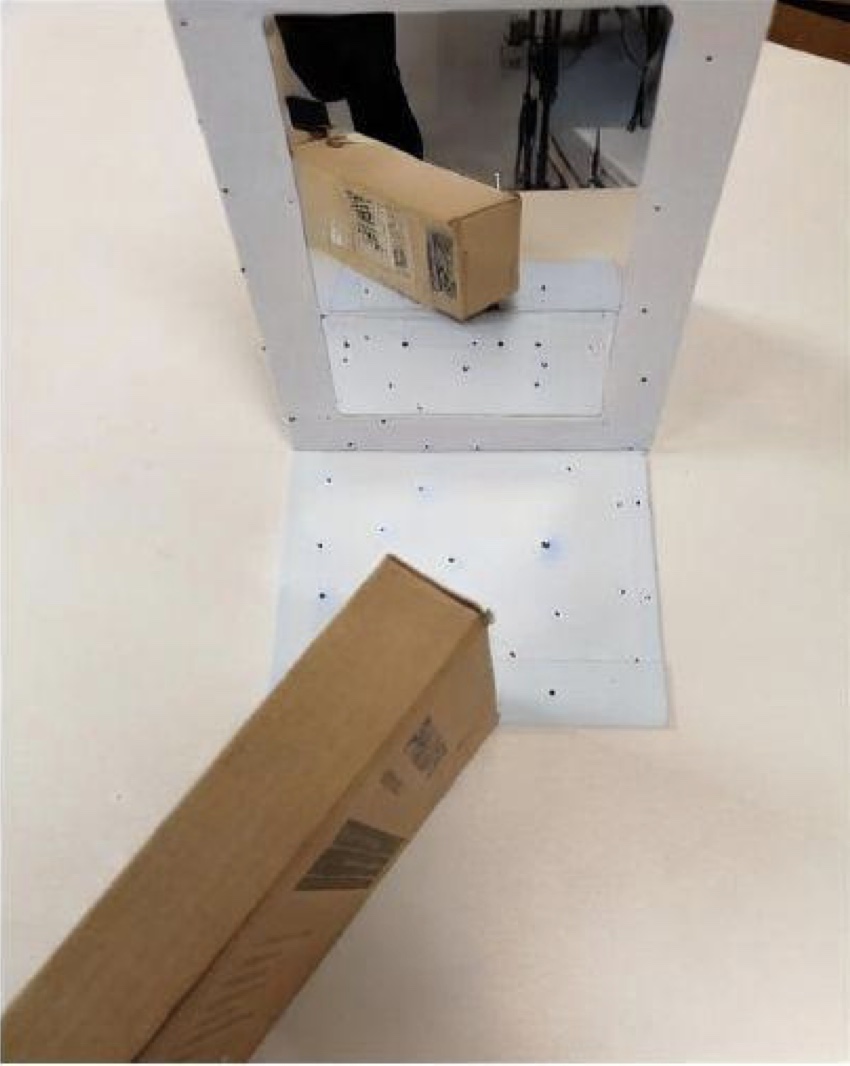}
\mirrorverserow{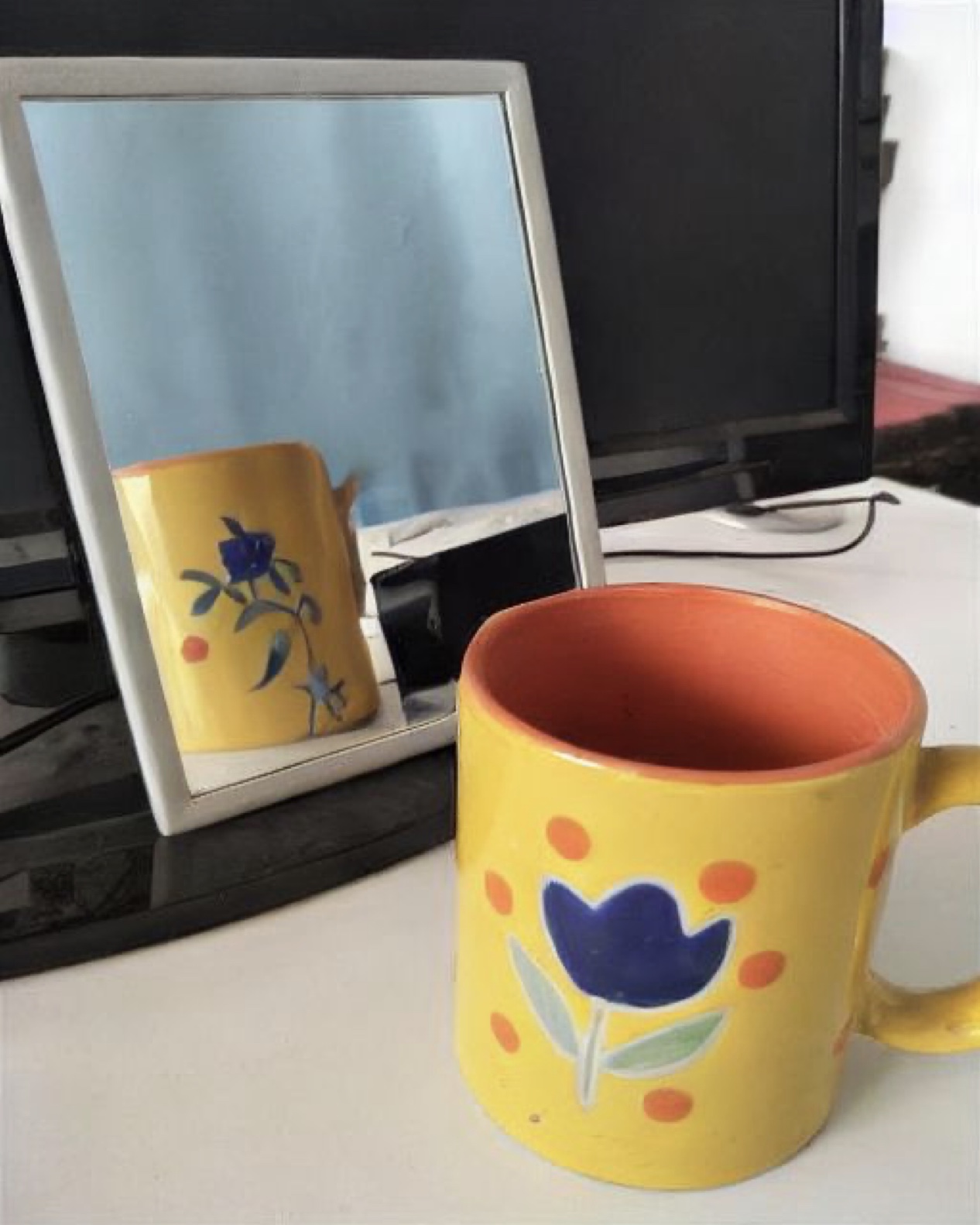}
\mirrorverserow{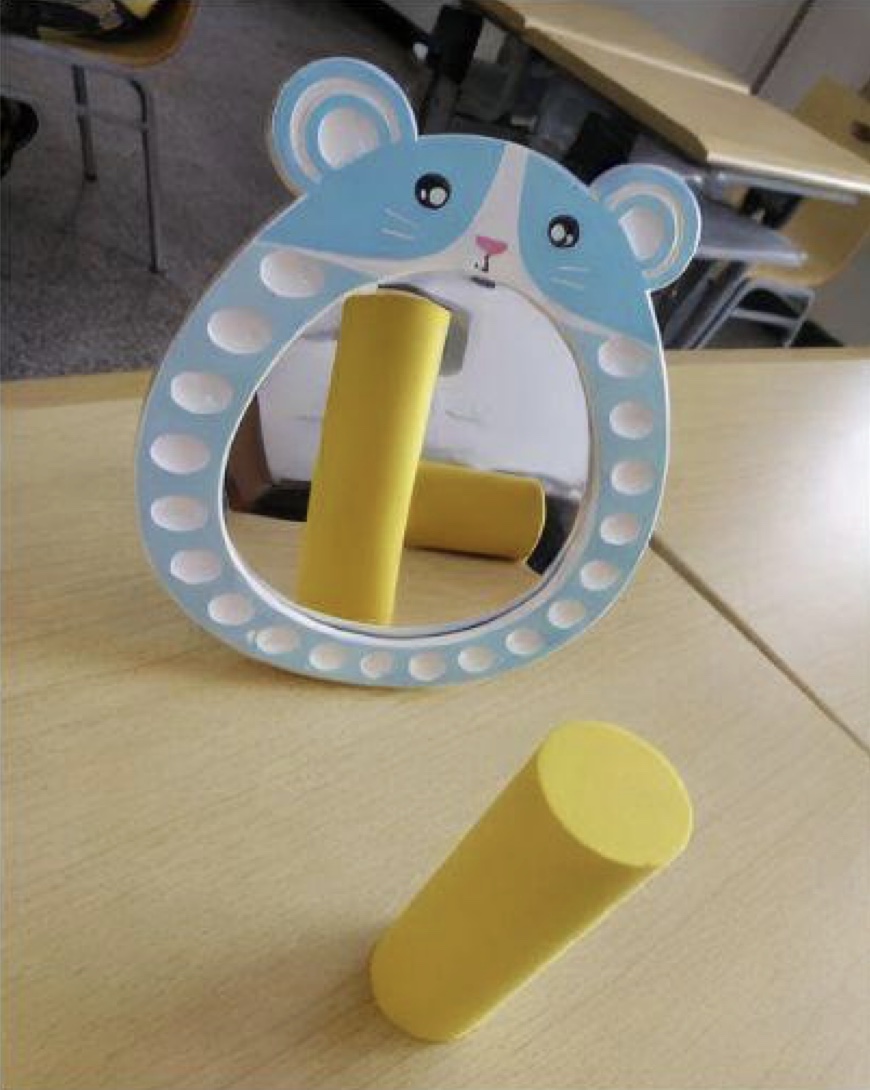}
\end{tabular}
\caption{\fromsupp{\textbf{Qualitative comparison on images from the MSD dataset.}
The examples, text prompts, and MirrorVerse outputs are taken directly from the MirrorVerse~\citep{dhiman2025mirrorverse} paper, rather than selected by us.
Columns (left to right): GT, MirrorVerse, FLUX.1 Fill, and Ours.
Even on these examples selected by the MirrorVerse authors, our method produces more geometrically consistent reflections, while the other methods exhibit incorrect orientation or content.}}
\label{fig:mirrorverse_examples}
\end{figure*}

\section{Noise Interpolation: Hyperparameter Selection and Qualitative Effect}
\label{sec:supp_hyperparam}

We analyze the effect of the noise interpolation parameters introduced in \cref{sec:inpainting}, which control the balance between geometric constraints and generative refinement. The exponent $\noiseInterpolationFunctionPower$ determines the sharpness of the transition between geometry-guided and generative noise predictions, while the start timestep $\firstTimestepToNoisePrediction$ defines the diffusion stage at which interpolation begins.

\subsection{Quantitative hyperparameter selection}
We quantitatively select $\noiseInterpolationFunctionPower$ and $\firstTimestepToNoisePrediction$ using the constrained-pixel masked PSNR. We first compare each configuration to the GT image and define a safe pool containing all configurations whose constrained-pixel masked PSNR is at most 0.5 dB below the best configuration. This threshold is used as a conservative heuristic for excluding configurations that deviate too much from the geometric constraints.

Among the configurations in this safe pool, we then choose the one with the lowest constrained-pixel masked PSNR against the initial projected image. A lower PSNR against the projection means that the output moved farther from the initial projection, which corresponds to more generative freedom. This two-stage rule therefore selects a configuration that stays close to the GT-constrained solution while maximizing generative refinement within the safe pool.

As discussed in \cref{sec:inpainting}, larger values of $\noiseInterpolationFunctionPower$ produce a sharper transition: the geometry-constraint mask's weight stays low for most of the interpolation interval, so the mask switches to the generative-refinement mask more abruptly near $\firstTimestepToNoisePrediction$.

\begin{figure*}[t]
  \centering
  \newcommand{\hyperpsnrimgwidth}{0.48\linewidth}
  \begin{minipage}{\hyperpsnrimgwidth}
    \centering
    \includegraphics[width=\linewidth]{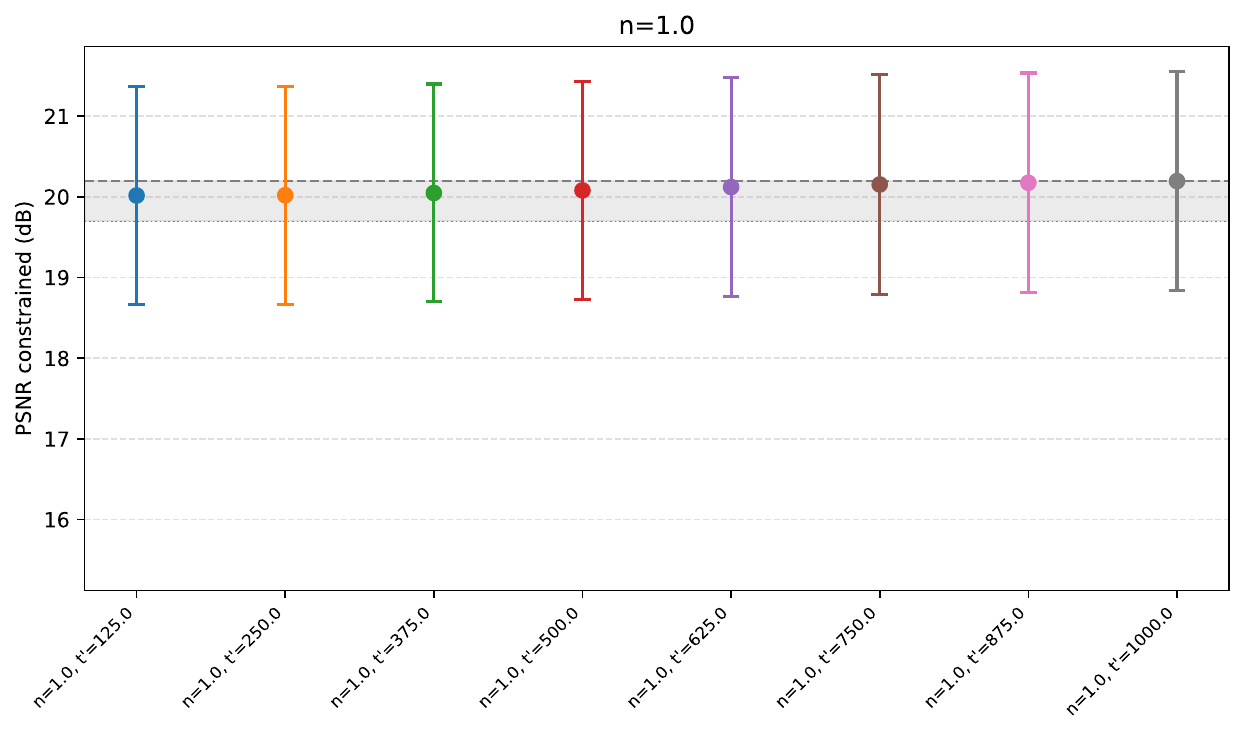}
  \end{minipage}\hfill
  \begin{minipage}{\hyperpsnrimgwidth}
    \centering
    \includegraphics[width=\linewidth]{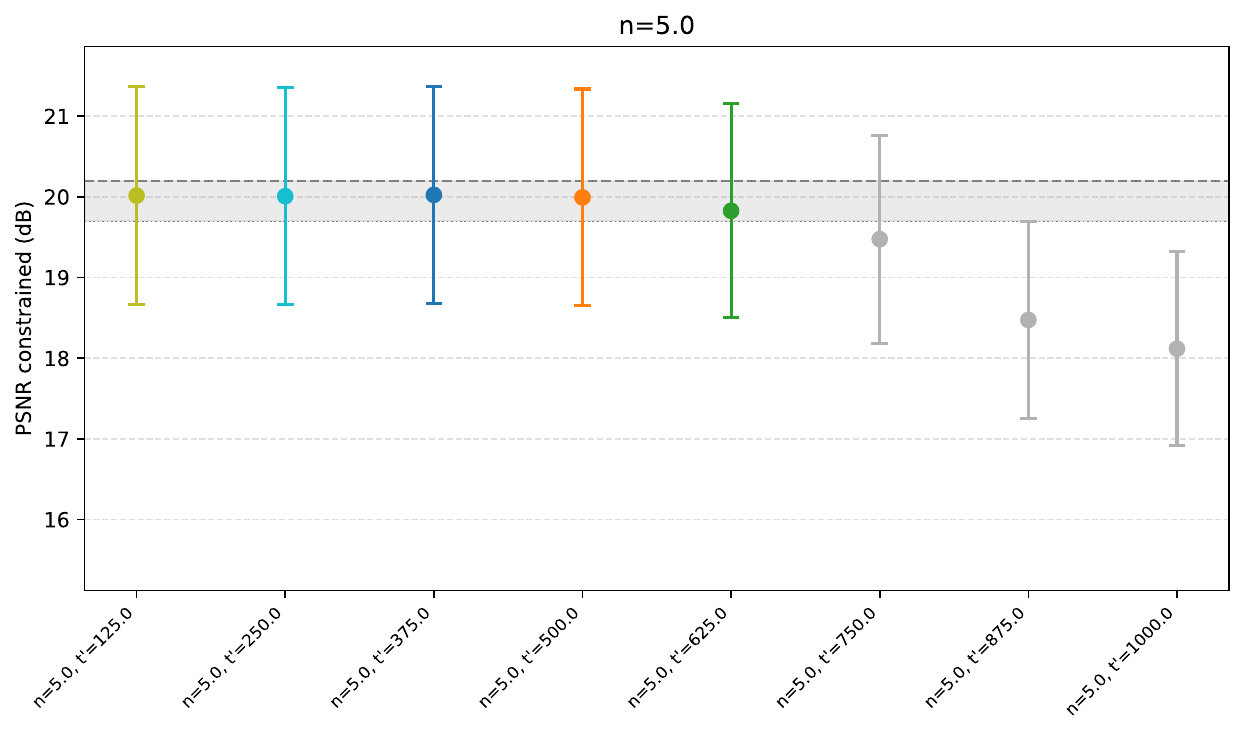}
  \end{minipage}

  \vspace{6pt}

  \begin{minipage}{\hyperpsnrimgwidth}
    \centering
    \includegraphics[width=\linewidth]{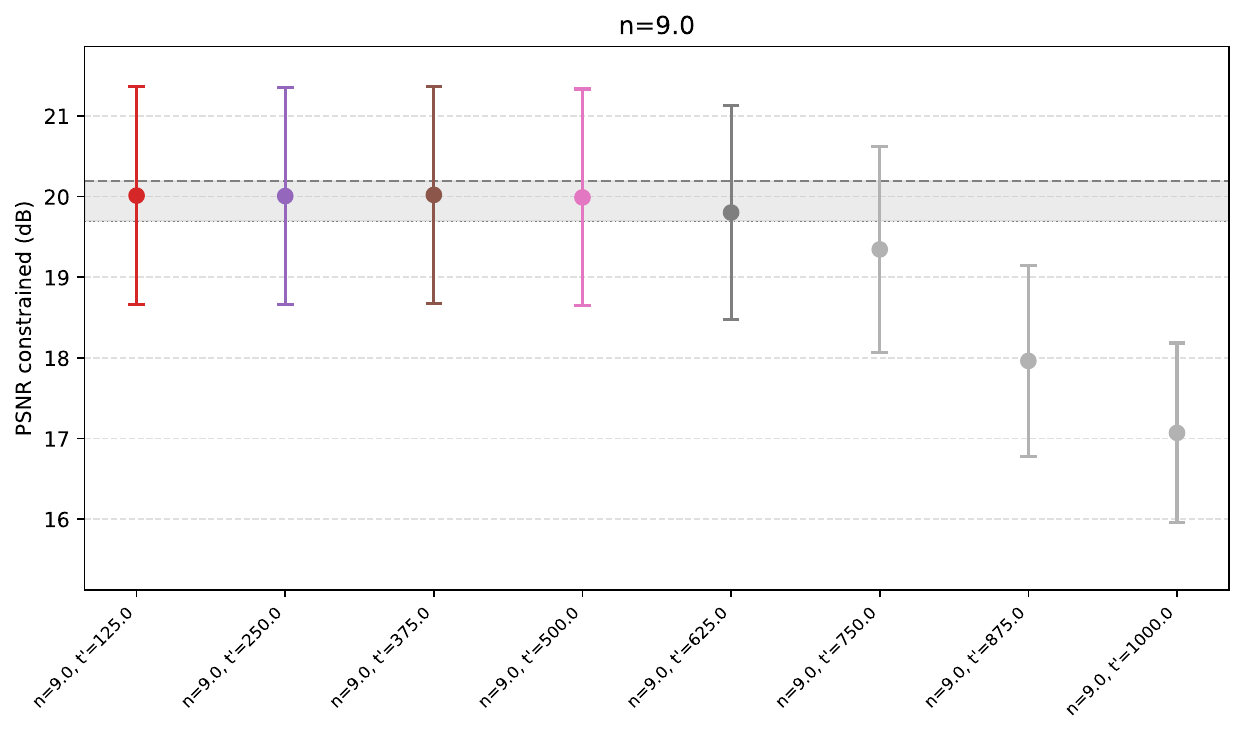}
  \end{minipage}\hfill
  \begin{minipage}{\hyperpsnrimgwidth}
    \centering
    \includegraphics[width=\linewidth]{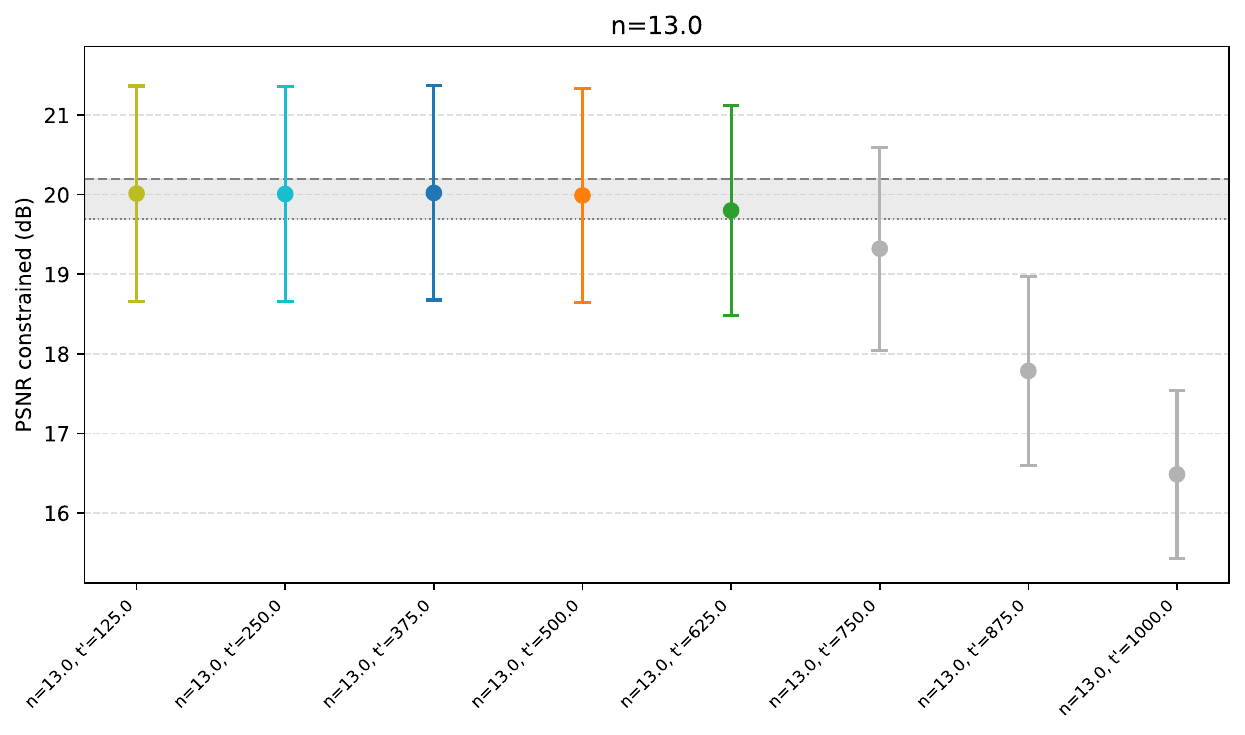}
  \end{minipage}
  \caption{\fromsupp{Constrained-pixel masked PSNR against the GT image for different $\noiseInterpolationFunctionPower$ and $\firstTimestepToNoisePrediction$ values. Gray configurations are more than 0.5 dB below the best GT masked PSNR and are therefore excluded from the safe pool; colored configurations remain inside the safe pool. Error bars show mean $\pm$ standard deviation across scenes. Earlier interpolation starts and sharper mask transitions increase generative freedom, but can also reduce adherence to the geometric constraints.}}
  \label{fig:hyperparam_selection_gt}
\end{figure*}

As shown in \cref{fig:hyperparam_selection_gt}, for $\noiseInterpolationFunctionPower=1$, the interpolation start time has little effect on the final result. For $\noiseInterpolationFunctionPower \in \{5,9,13\}$, the interpolation has a stronger effect when it begins earlier. In general, earlier interpolation starts and sharper mask transitions provide more generative freedom, but can also make the model ignore the geometric constraints. This effect becomes stronger for $\noiseInterpolationFunctionPower \geq 5$ and $\firstTimestepToNoisePrediction \geq 750$. In contrast, for $\noiseInterpolationFunctionPower \geq 5$ and $\firstTimestepToNoisePrediction \leq 625$, the differences in constrained-pixel masked PSNR against the GT image are small. When $\noiseInterpolationFunctionPower \geq 5$ and $\firstTimestepToNoisePrediction > 625$, the choice of $\noiseInterpolationFunctionPower$ becomes more important.

\begin{figure*}[t]
  \centering
  \newcommand{\hyperpsnrimgwidth}{0.48\linewidth}
  \begin{minipage}{\hyperpsnrimgwidth}
    \centering
    \includegraphics[width=\linewidth]{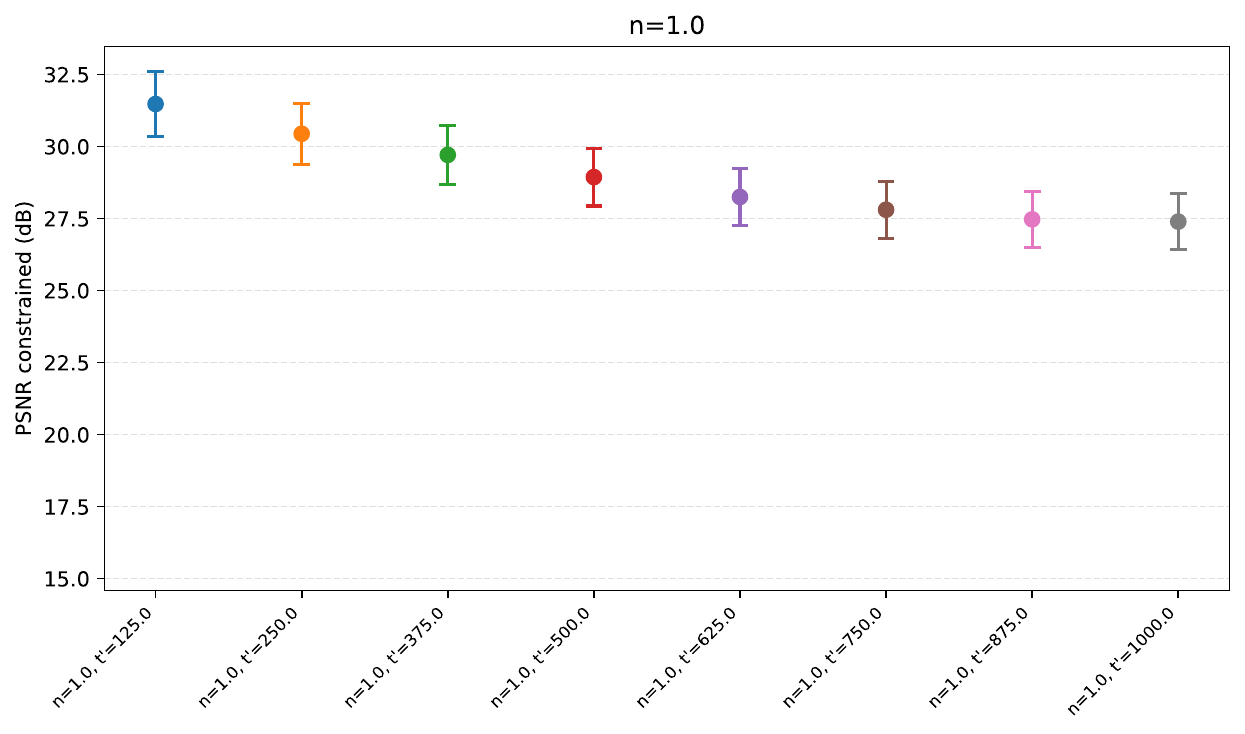}
  \end{minipage}\hfill
  \begin{minipage}{\hyperpsnrimgwidth}
    \centering
    \includegraphics[width=\linewidth]{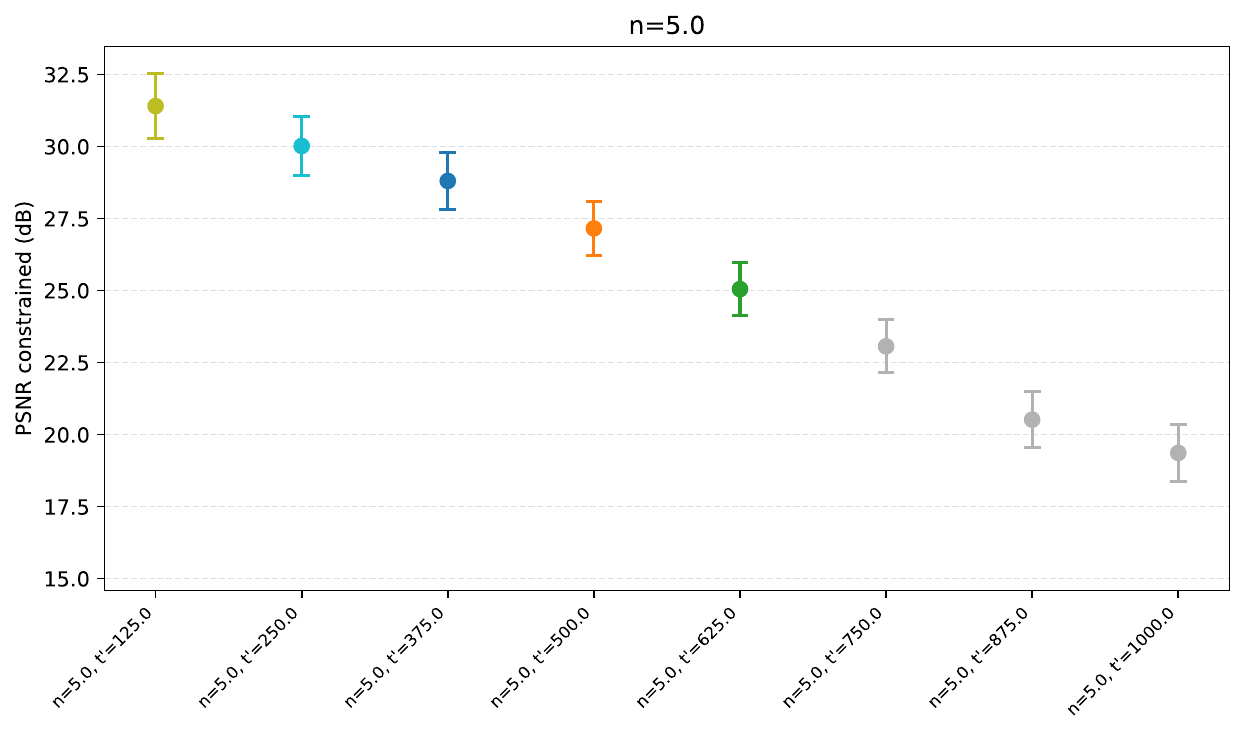}
  \end{minipage}

  \vspace{6pt}

  \begin{minipage}{\hyperpsnrimgwidth}
    \centering
    \includegraphics[width=\linewidth]{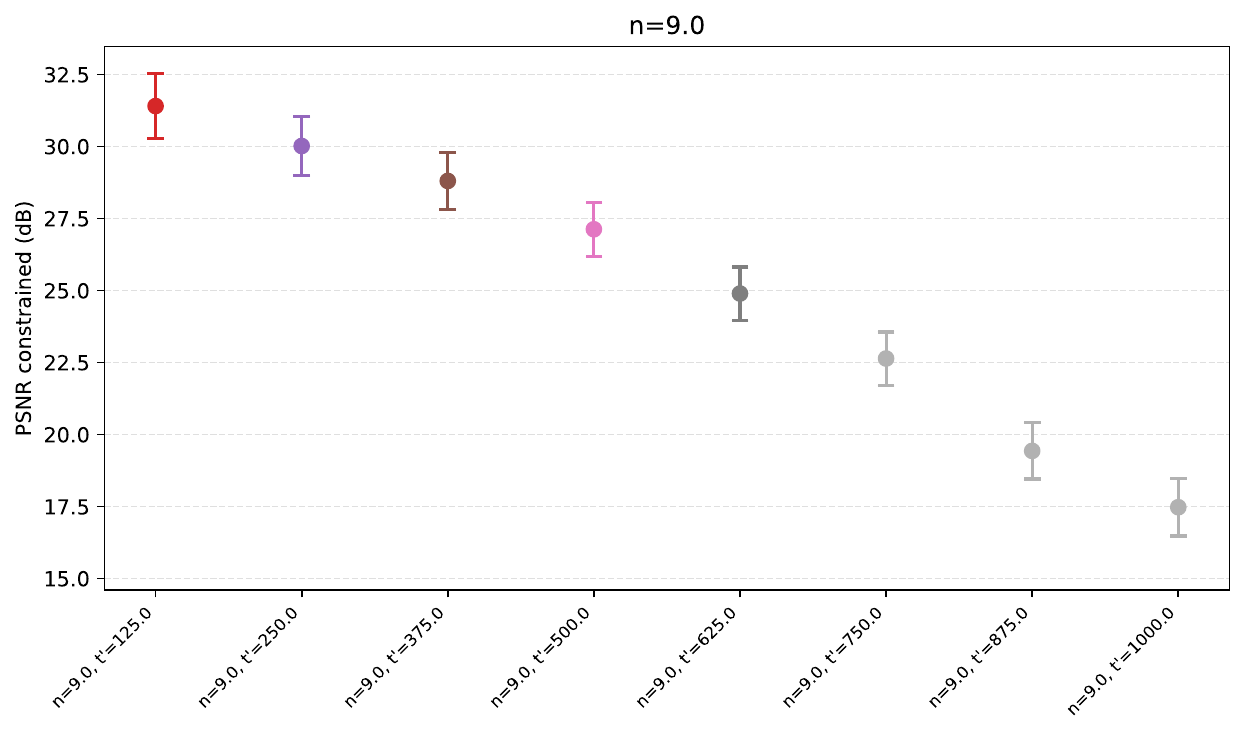}
  \end{minipage}\hfill
  \begin{minipage}{\hyperpsnrimgwidth}
    \centering
    \includegraphics[width=\linewidth]{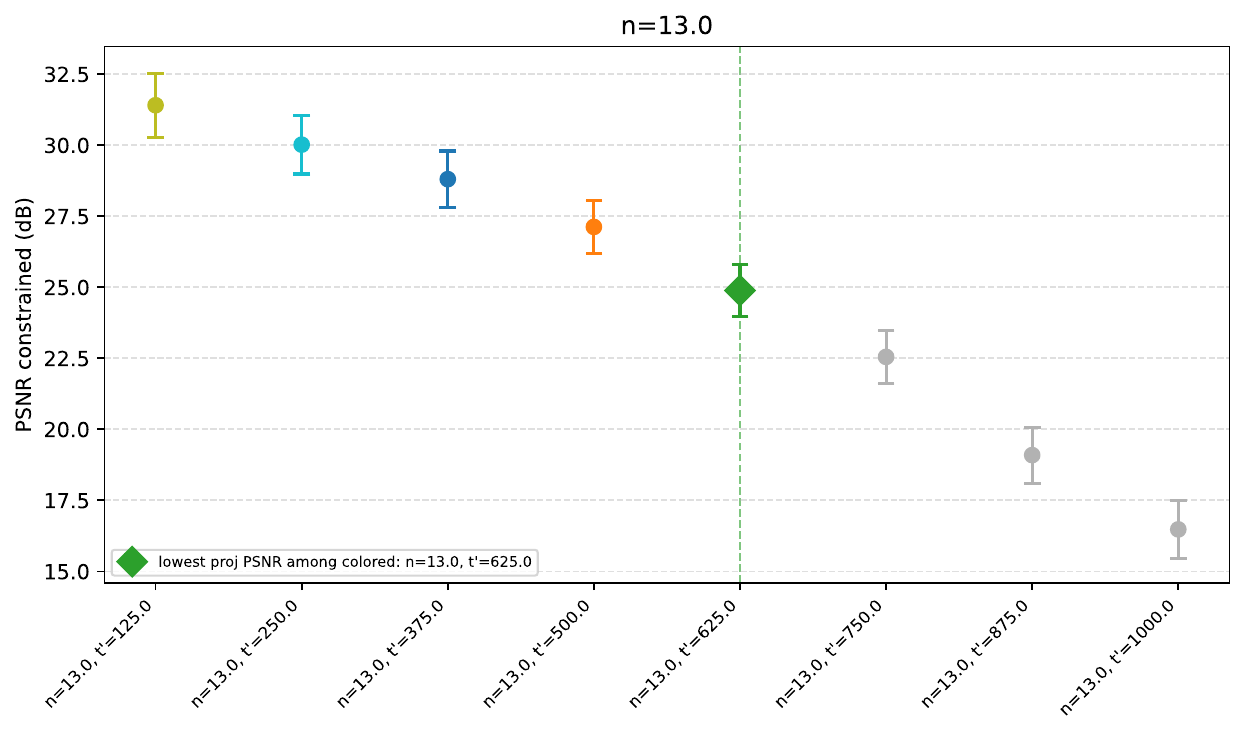}
  \end{minipage}
  \caption{\fromsupp{Constrained-pixel masked PSNR against the initial projected image. The colors match the safe-pool configurations from the GT comparison, while gray configurations are outside the safe pool. Circles denote non-selected configurations, and the rhombus marks the selected configuration. The selected configuration is the safe-pool member with the lowest projected-image PSNR, indicating the largest change from the projection while remaining within the GT-based safe pool. Error bars show mean $\pm$ standard deviation across scenes.}}
  \label{fig:hyperparam_selection_projected}
\end{figure*}

As shown in \cref{fig:hyperparam_selection_projected}, the projected-image comparison shows a similar trend: earlier interpolation and sharper transitions lead to a larger departure from the projection. We therefore choose the configuration that maximizes this departure from the projection among the configurations that remain within 0.5 dB of the best GT masked PSNR. Qualitative examples illustrating the effect of this choice are shown in \cref{fig:hyperparam_nt_qualitative_1,fig:hyperparam_nt_qualitative_2}.

\newcommand{\zoomSrcX}{-0.4}          
\newcommand{\zoomSrcY}{-0.16}          
\newcommand{\zoomMagnification}{2.6}   
\newcommand{\zoomDstSizeFrac}{0.46}    
\newcommand{\zoomDstX}{0.24}           
\newcommand{\zoomDstY}{0.24}           

\newlength{\hyperntimgwidth}
\setlength{\hyperntimgwidth}{0.21\textwidth}
\newlength{\zoomDstLen}
\setlength{\zoomDstLen}{\zoomDstSizeFrac\hyperntimgwidth}

\newcommand{\zoomimg}[1]{%
  \begin{tikzpicture}[baseline,
      x=\hyperntimgwidth, y=\hyperntimgwidth,
      spy using outlines={rectangle,
        magnification=\zoomMagnification,
        width=\zoomDstLen, height=\zoomDstLen,
        connect spies}]
    \node[inner sep=0pt] {\includegraphics[width=\hyperntimgwidth]{#1}};
    \spy [red] on (\zoomSrcX,\zoomSrcY) in node[inner sep=0pt] at (\zoomDstX,\zoomDstY);
  \end{tikzpicture}%
}

\newcommand{\hyperparamntrow}[2]{%
  \zoomimg{imgs/hyperparam_nt/rank#1_n#2_t125.0.jpg}&
  \zoomimg{imgs/hyperparam_nt/rank#1_n#2_t375.0.jpg}&
  \zoomimg{imgs/hyperparam_nt/rank#1_n#2_t625.0.jpg}&
  \zoomimg{imgs/hyperparam_nt/rank#1_n#2_t1000.0.jpg}%
}

\newcommand{\hyperparamntrowmarked}[2]{%
  \zoomimg{imgs/hyperparam_nt/rank#1_n#2_t125.0.jpg}&
  \zoomimg{imgs/hyperparam_nt/rank#1_n#2_t375.0.jpg}&
  {\setlength{\fboxsep}{0pt}\setlength{\fboxrule}{2pt}\fcolorbox{blue}{white}{\zoomimg{imgs/hyperparam_nt/rank#1_n#2_t625.0.jpg}}}&
  \zoomimg{imgs/hyperparam_nt/rank#1_n#2_t1000.0.jpg}%
}

\newcommand{\hyperparamntrowplain}[2]{%
  \includegraphics[width=\hyperntimgwidth]{imgs/hyperparam_nt/rank#1_n#2_t125.0.jpg}&
  \includegraphics[width=\hyperntimgwidth]{imgs/hyperparam_nt/rank#1_n#2_t375.0.jpg}&
  \includegraphics[width=\hyperntimgwidth]{imgs/hyperparam_nt/rank#1_n#2_t625.0.jpg}&
  \includegraphics[width=\hyperntimgwidth]{imgs/hyperparam_nt/rank#1_n#2_t1000.0.jpg}%
}
\newcommand{\hyperparamntrowplainmarked}[2]{%
  \includegraphics[width=\hyperntimgwidth]{imgs/hyperparam_nt/rank#1_n#2_t125.0.jpg}&
  \includegraphics[width=\hyperntimgwidth]{imgs/hyperparam_nt/rank#1_n#2_t375.0.jpg}&
  {\setlength{\fboxsep}{0pt}\setlength{\fboxrule}{2pt}\fcolorbox{blue}{white}{\includegraphics[width=\hyperntimgwidth]{imgs/hyperparam_nt/rank#1_n#2_t625.0.jpg}}}&
  \includegraphics[width=\hyperntimgwidth]{imgs/hyperparam_nt/rank#1_n#2_t1000.0.jpg}%
}

\begin{figure*}[t]
  \centering
  \setlength{\tabcolsep}{1.5pt}
  \renewcommand{\arraystretch}{0.6}
  \begin{tabular}{@{}>{\raggedleft\arraybackslash}m{2cm}@{\hspace{4pt}}cccc@{}}
    \small $\firstTimestepToNoisePrediction\!\rightarrow$ &
    \small $125$ & \small $375$ & \small $625$ & \small $1000$ \\[3pt]
    \small $\noiseInterpolationFunctionPower{=}1$  & \hyperparamntrow{11}{1.0}       \\[3pt]
    \small $\noiseInterpolationFunctionPower{=}5$  & \hyperparamntrow{11}{5.0}       \\[3pt]
    \small $\noiseInterpolationFunctionPower{=}9$  & \hyperparamntrow{11}{9.0}       \\[3pt]
    \small $\noiseInterpolationFunctionPower{=}13$ & \hyperparamntrowmarked{11}{13.0} \\
  \end{tabular}
  \caption{%
    \fromsupp{Qualitative influence of $\noiseInterpolationFunctionPower$ (rows) and
    $\firstTimestepToNoisePrediction$ (columns) on the generated image for Example~1.
    When $\firstTimestepToNoisePrediction \leq 375$ or when $\noiseInterpolationFunctionPower{=}1$,
    smearing artifacts in the pillow, visible on the left side of the mirror, are not corrected
    by the generative refinement mask.
    Conversely, at $\noiseInterpolationFunctionPower{=}13$, $\firstTimestepToNoisePrediction{=}1000$,
    the model increasingly ignores the projected content and relies on its learned prior,
    producing incorrect shadow reflections and wrong sofa orientation.
    The blue outline marks our chosen configuration
    ($\noiseInterpolationFunctionPower{=}13$, $\firstTimestepToNoisePrediction{=}625$),
    which avoids both failure modes.}%
  }
  \label{fig:hyperparam_nt_qualitative_1}
\end{figure*}

\begin{figure*}[t]
  \centering
  \setlength{\tabcolsep}{1.5pt}
  \renewcommand{\arraystretch}{0.6}
  \begin{tabular}{@{}>{\raggedleft\arraybackslash}m{2cm}@{\hspace{4pt}}cccc@{}}
    \small $\firstTimestepToNoisePrediction\!\rightarrow$ &
    \small $125$ & \small $375$ & \small $625$ & \small $1000$ \\[3pt]
    \small $\noiseInterpolationFunctionPower{=}1$  & \hyperparamntrowplain{4}{1.0}       \\[3pt]
    \small $\noiseInterpolationFunctionPower{=}5$  & \hyperparamntrowplain{4}{5.0}       \\[3pt]
    \small $\noiseInterpolationFunctionPower{=}9$  & \hyperparamntrowplain{4}{9.0}       \\[3pt]
    \small $\noiseInterpolationFunctionPower{=}13$ & \hyperparamntrowplainmarked{4}{13.0} \\
  \end{tabular}
  \caption{%
    \fromsupp{Qualitative influence of $\noiseInterpolationFunctionPower$ (rows) and
    $\firstTimestepToNoisePrediction$ (columns) on the generated image for Example~2.
    At $\firstTimestepToNoisePrediction{=}1000$ with $\noiseInterpolationFunctionPower{\geq}5$,
    the model fails to preserve the sofa reflection; for $\noiseInterpolationFunctionPower{\geq}9$,
    it additionally shrinks the mirror region.
    The blue outline marks our chosen configuration
    ($\noiseInterpolationFunctionPower{=}13$, $\firstTimestepToNoisePrediction{=}625$).}%
  }
  \label{fig:hyperparam_nt_qualitative_2}
\end{figure*}

\section{RCS Parameter Selection and Validation}
\label{sec:supp_rcs_sweep}

This section provides the parameter-selection and validation details omitted from the main text.

We select the dilation radius $r$ and iteration count $i$ for the Reflection Consistency Score (RCS) mask estimation by performing a grid search over both parameters on 12 of the 15 Blender scenes in our hyperparameter validation set (resolution $800{\times}800$), excluding 3 scenes with fewer than 500 matched correspondence pixels as their sparsity is unrepresentative of the dataset. For each candidate pair $(r,i)$, we apply the dilation step described in \cref{sec:reflection_consistency_score}, intersect the result with the mirror region, and compare against the ground-truth constraint mask using F$_\beta$ with $\beta{=}0.5$, a criterion that weights precision $4\times$ over recall, favouring mask fidelity while still penalising overly sparse masks.

\begin{figure*}[htbp]
  \centering
  \newcommand{\fbetaimgwidth}{0.16\linewidth}
  \begin{minipage}{\fbetaimgwidth}
    \centering
    \small $r=1$\\[2pt]
    \includegraphics[width=\linewidth]{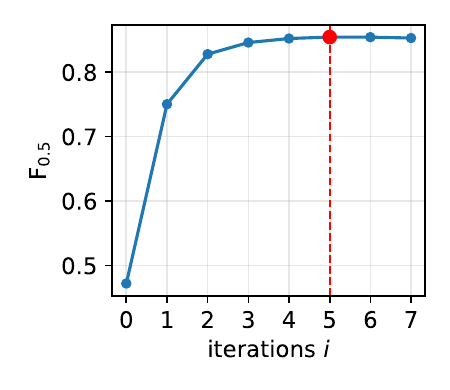}
  \end{minipage}\hfill
  \begin{minipage}{\fbetaimgwidth}
    \centering
    \small $r=2$\\[2pt]
    \includegraphics[width=\linewidth]{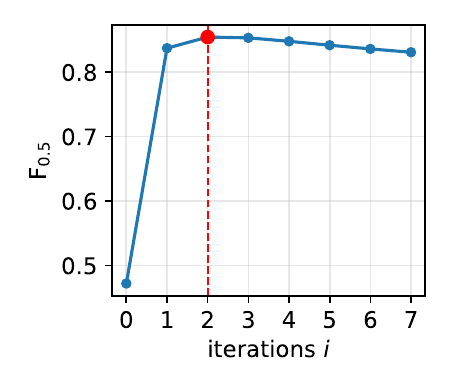}
  \end{minipage}\hfill
  \begin{minipage}{\fbetaimgwidth}
    \centering
    \small $r=3$\\[2pt]
    \includegraphics[width=\linewidth]{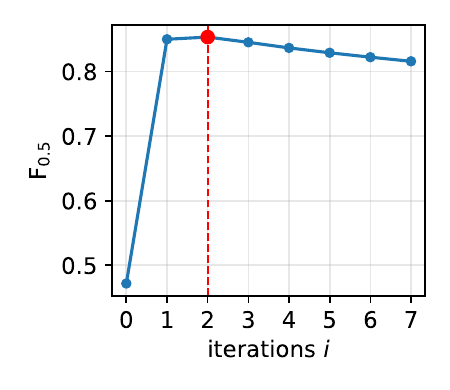}
  \end{minipage}\hfill
  \begin{minipage}{\fbetaimgwidth}
    \centering
    \small $r=4$\\[2pt]
    \includegraphics[width=\linewidth]{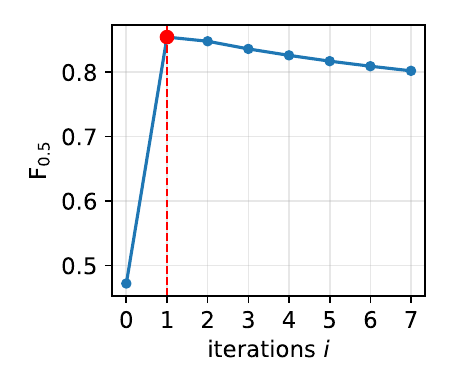}
  \end{minipage}\hfill
  \begin{minipage}{\fbetaimgwidth}
    \centering
    \small $r=5$\\[2pt]
    \includegraphics[width=\linewidth]{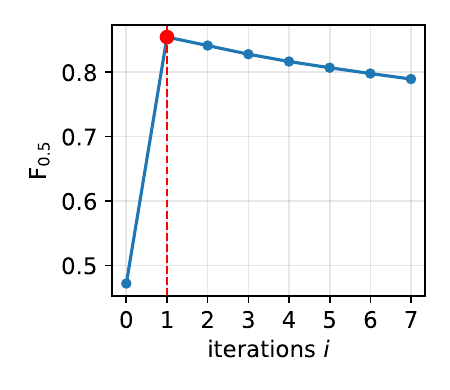}
  \end{minipage}\hfill
  \begin{minipage}{\fbetaimgwidth}
    \centering
    \small $r=6$\\[2pt]
    \includegraphics[width=\linewidth]{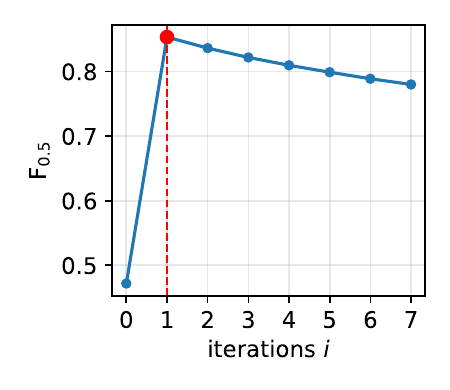}
  \end{minipage}
  \caption{%
    \fromsupp{F$_{0.5}$ score versus dilation iteration count $i$, for each dilation radius $r \in \{1,\ldots,6\}$
    (mean across the 12 Blender scenes with at least 500 matched correspondence pixels;
    3 scenes with sparser correspondences are excluded as unrepresentative of the dataset;
    $r{=}0$, i.e.\ no dilation, is omitted as it tops out far below $r{\geq}1$).
    The red dashed vertical line marks the iteration count that maximises F$_{0.5}$ for that radius;
    the red dot marks the corresponding F$_{0.5}$ value.
    Because $\beta{=}0.5$ weights precision $4\times$ over recall,
    the criterion favours mask fidelity while still requiring reasonable coverage.
    The optimal iteration count varies with radius ($r{=}1{:}~i{=}5$; $r{=}2,3{:}~i{=}2$;
    $r{=}4,5,6{:}~i{=}1$), with $r{=}5$ achieving the
    best overall score (F$_{0.5}{=}0.85$, precision~$0.85$, recall~$0.90$).}
  }
  \label{fig:rcs_fbeta_plots}
\end{figure*}

\Cref{fig:rcs_fbeta_plots} shows the mean F$_{0.5}$ score as the iteration count varies, with one panel per radius $r \in \{1,\ldots,6\}$. Few iterations yield sparse masks with high precision but low recall; more iterations expand coverage at the cost of precision. The F$_{0.5}$ criterion selects \textbf{radius~$=5$, iterations~$=1$}, achieving mean precision~$0.85$, recall~$0.90$, and F$_{0.5}{=}0.85$, which we use for an $800{\times}800$ image throughout.

%
%

\Cref{fig:rcs_dilation_grid} shows per-iteration overlays for two representative Blender scenes alongside the ground-truth constraint mask, illustrating how additional dilation iterations progressively expand mask coverage at the selected radius.

%
%

\newcommand{\rcsdilationimgwidth}{0.19\textwidth}

\newcommand{\rcsdilationcell}[2]{%
  \includegraphics[width=\rcsdilationimgwidth]{%
    imgs/rcs_dilation_sweep/sample_#1/overlay_r5_i#2%
  }%
}
\newcommand{\rcsdilationgt}[1]{%
  \includegraphics[width=\rcsdilationimgwidth]{%
    imgs/rcs_dilation_sweep/sample_#1/overlay_gt%
  }%
}

\begin{figure*}[htbp]
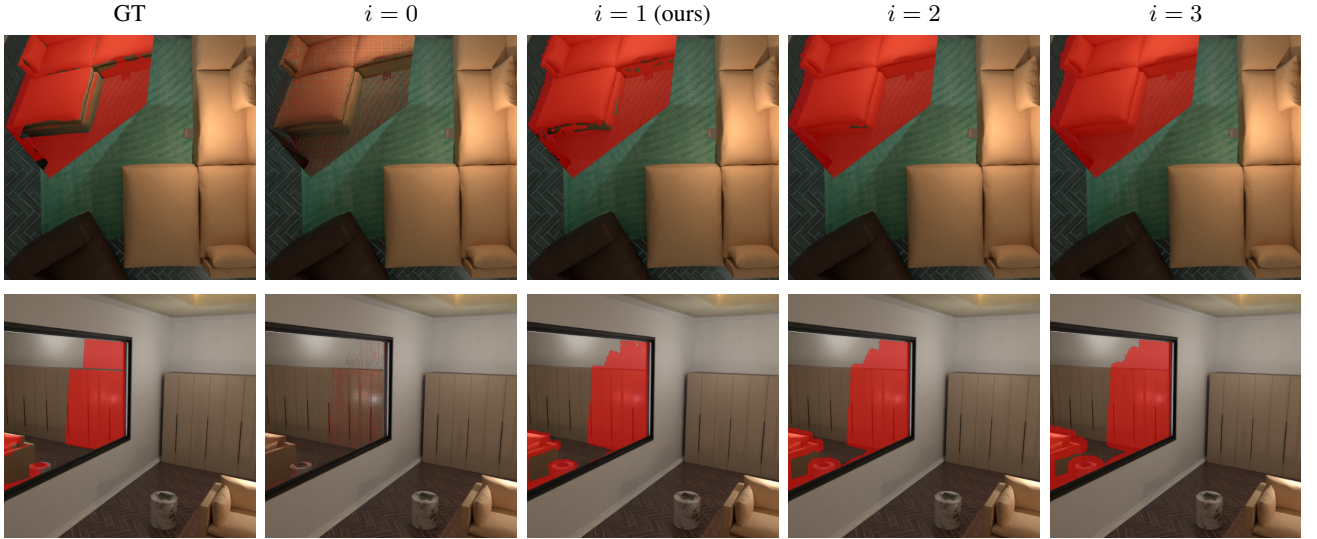

  \centering
  \setlength{\tabcolsep}{2pt}
  \begin{tabular}{*{5}{c}}
    \small GT &
    \small $i=0$ &
    \small $i=1$ (ours) &
    \small $i=2$ &
    \small $i=3$ \\[2pt]
    \rcsdilationgt{10} &
    \rcsdilationcell{10}{0} &
    \rcsdilationcell{10}{1} &
    \rcsdilationcell{10}{2} &
    \rcsdilationcell{10}{3} \\[2pt]
    \rcsdilationgt{12} &
    \rcsdilationcell{12}{0} &
    \rcsdilationcell{12}{1} &
    \rcsdilationcell{12}{2} &
    \rcsdilationcell{12}{3} \\[2pt]
  \end{tabular}
  \caption{%
    \fromsupp{RCS dilation-iteration sweep (radius $r{=}5$, the selected operating radius)
    for two Blender scenes.
    The first column shows the ground-truth (GT) geometrically-constrained mirror pixels.
    The remaining columns show the estimated RCS mask (red overlay) for increasing iteration
    counts $i \in \{0,\ldots,3\}$; the selected operating point $i{=}1$ is labelled ``ours''.
    Fewer iterations ($i{=}0$) leave the mask overly sparse, while more iterations ($i{=}2,3$)
    over-expand it beyond the true constrained region; the selected $i{=}1$ visibly gives the
    best balance between precision and recall.}
  }
  \label{fig:rcs_dilation_grid}
\end{figure*}

\begin{figure*}[h]
  \centering
  \includegraphics[width=0.55\linewidth]{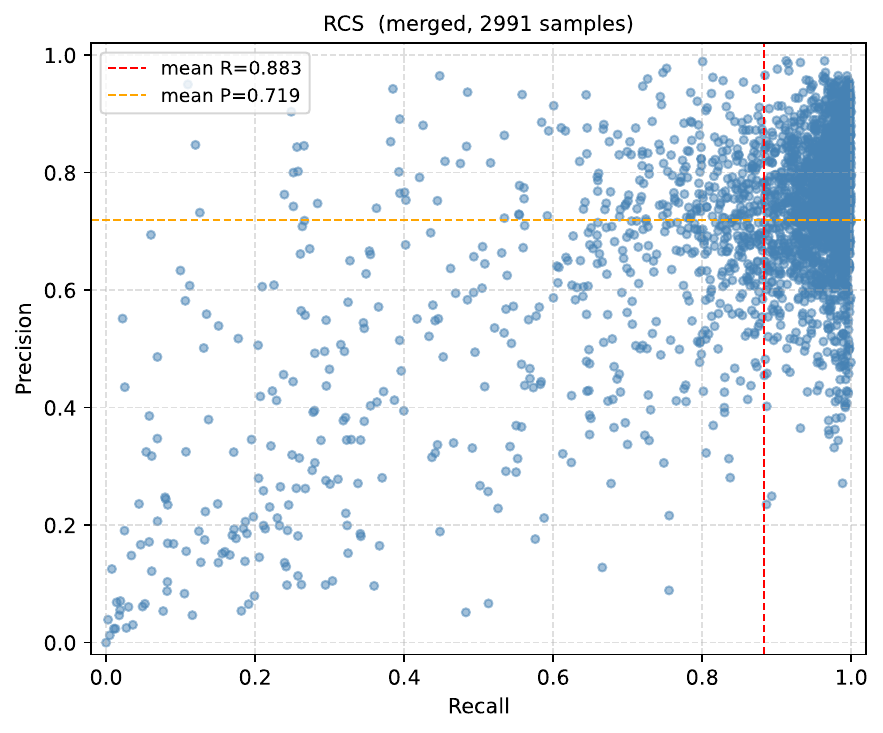}
  \caption{%
    \new{Per-scene precision and recall of the RCS mask against the ground-truth
    geometry-constrained mask, across all the scenes in MirrorBench-V2
    (chosen operating point: radius~$=5$, iterations~$=1$).
    The RCS mask achieves a mean precision of $0.72$ ($\sigma{=}0.16$) and a
    mean recall of $0.88$ ($\sigma{=}0.20$), indicating that it reliably
    approximates the ground-truth constrained pixels across scenes.}
  }
  \label{fig:rcs_precision_recall_scatter}
\end{figure*}

\Cref{fig:rcs_precision_recall_scatter} shows the per-scene precision and recall of the estimated RCS masks against the MirrorBench-V2 ground-truth constraint masks, underlying the aggregate validation results reported in \cref{sec:reflection_consistency_score}.

\section{Seed Consistency}
\label{sec:supp_seed_variance_full_mirror}

\begin{figure*}[htbp]
  \centering
  \begin{subfigure}[b]{0.49\linewidth}
    \includegraphics[width=\linewidth]{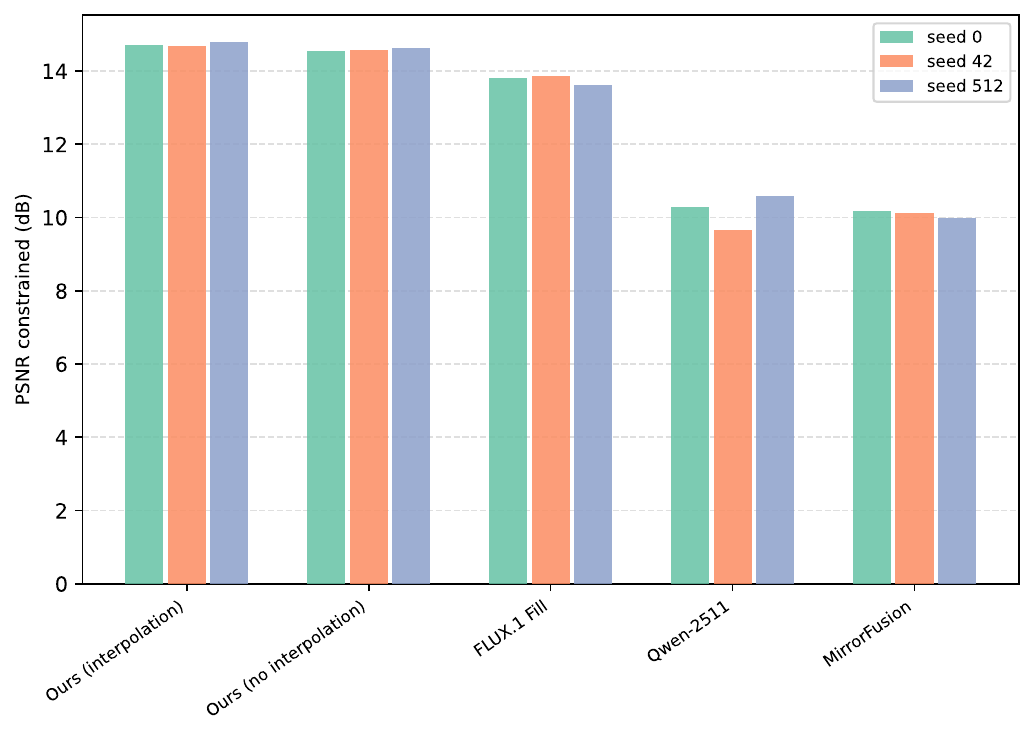}
  \end{subfigure}
  \hfill
  \begin{subfigure}[b]{0.49\linewidth}
    \includegraphics[width=\linewidth]{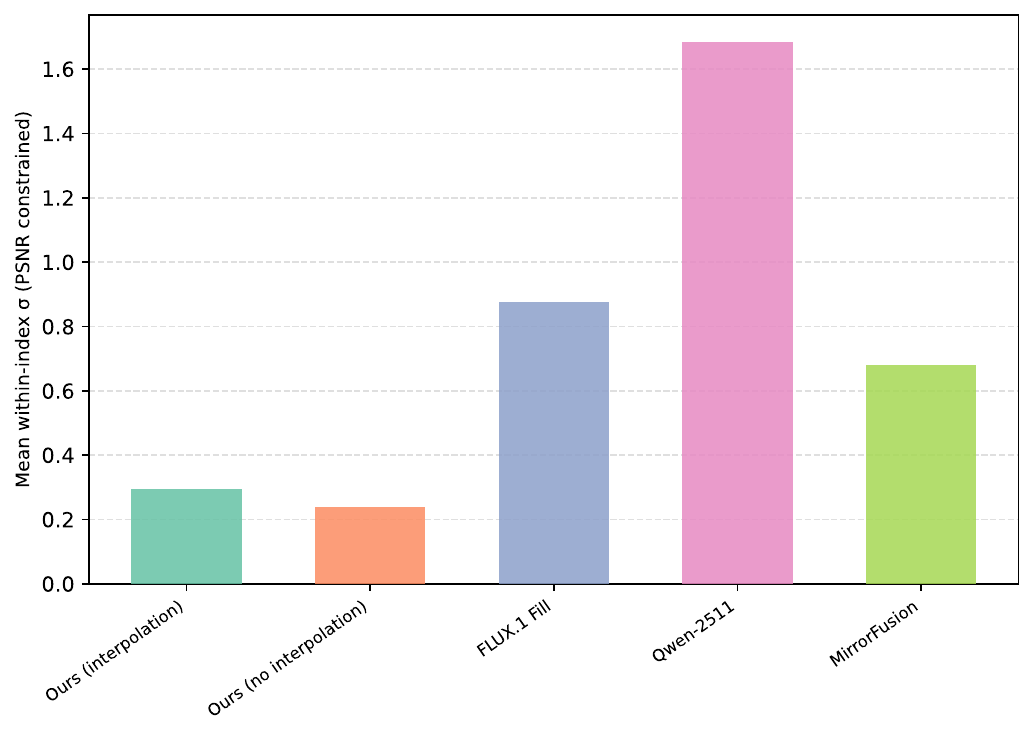}
  \end{subfigure}
  \caption{
    Seed consistency on real captured images over the geometrically constrained region.
    \textbf{Left:} Average PSNR per method across three random seeds.
    \textbf{Right:} Mean within-image standard deviation across seeds (lower is better); a taller bar means the method is more sensitive to the choice of random seed.
    Our method produces substantially more consistent results than the baselines, indicating that our geometric conditioning anchors the output to the scene's physical constraints rather than relying on random generation.
    Interpolation slightly raises the deviation, reflecting that the refinement mask genuinely refines the overall structure --- including the constrained pixels --- rather than leaving them fixed.
  }
  \label{fig:seed_variance_psnr}
\end{figure*}

\Cref{fig:seed_variance_psnr} analyzes seed consistency on real images over the geometrically constrained region. Our method not only outperforms the baselines but also produces substantially more stable outputs across seeds, suggesting that our geometric conditioning anchors the output to physical scene constraints rather than relying on random generation.

\begin{figure*}[t]
  \centering
  \begin{minipage}[b]{0.49\linewidth}
    \centering
    \includegraphics[width=\linewidth]{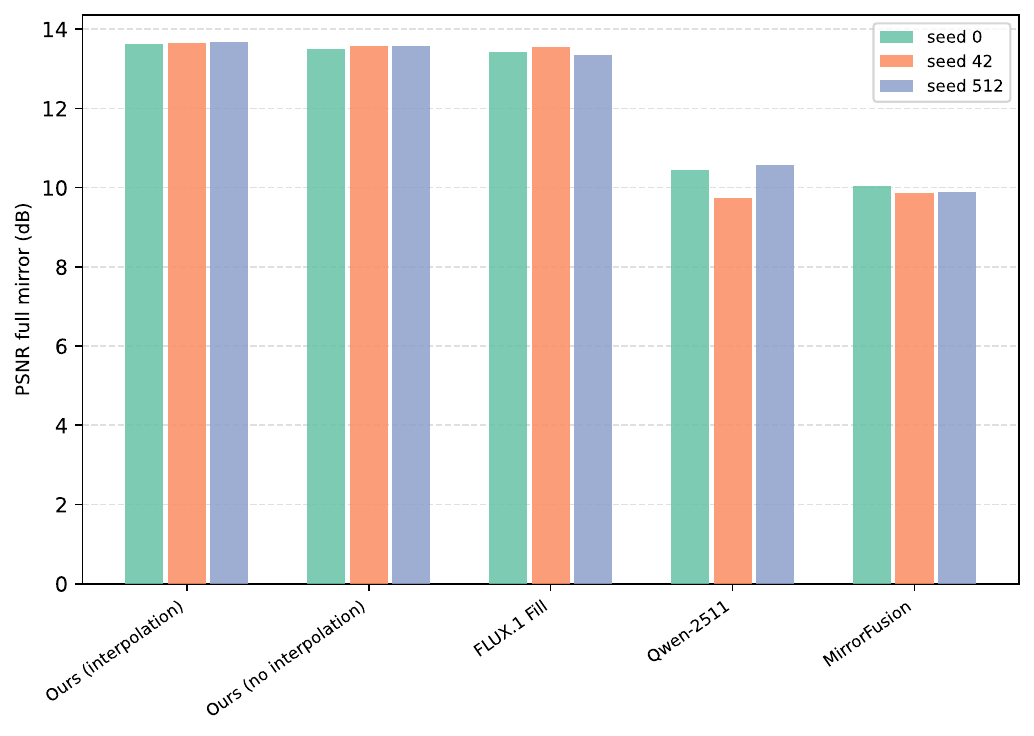}
  \end{minipage}\hfill
  \begin{minipage}[b]{0.49\linewidth}
    \centering
    \includegraphics[width=\linewidth]{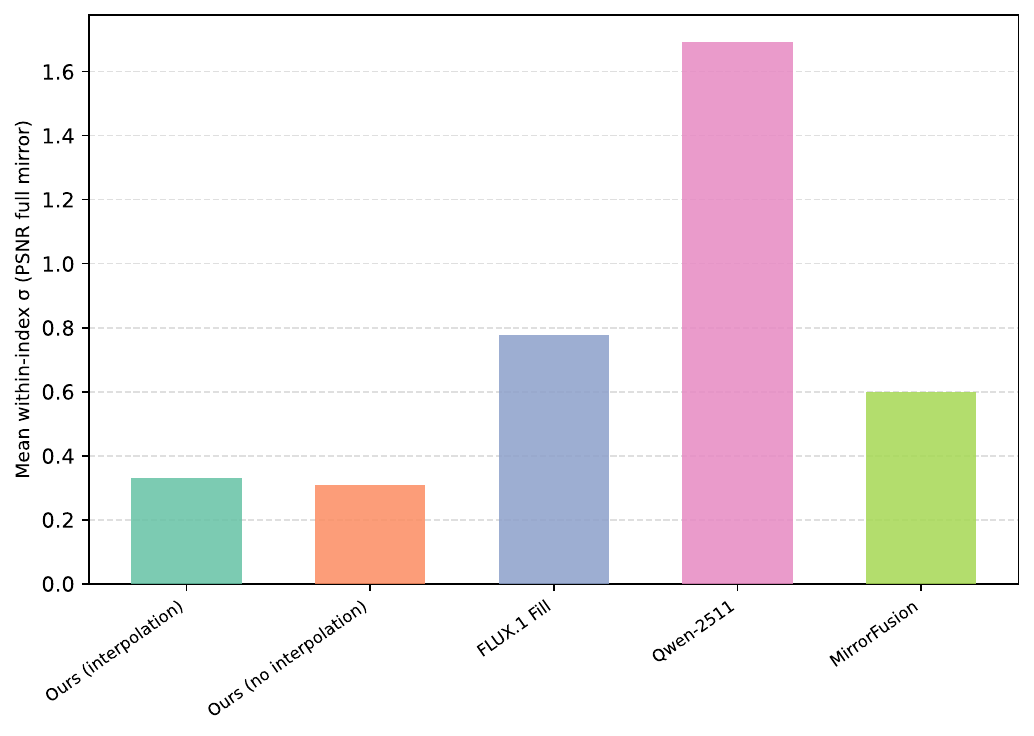}
  \end{minipage}
  \caption{
    \fromsupp{Seed consistency analysis on real captured images (PSNR over the full mirror mask).
    \textbf{Left:} Average PSNR per method, broken down by random seed.
    Each group of bars shows the same model evaluated with three different random seeds.
    \textbf{Right:} Mean within-image standard deviation across seeds per method (lower is better).
    Compared to the constrained-region results in \cref{fig:seed_variance_psnr}, the variance gap between methods is smaller here, as unconstrained mirror pixels can be plausibly hallucinated in multiple ways by all methods.}
  }
  \label{fig:seed_variance_psnr_full_mirror}
\end{figure*}

\Cref{fig:seed_variance_psnr_full_mirror} shows the corresponding analysis over the full mirror mask.

As expected, this variance gap is smaller when measured over the full mirror mask. Unlike the geometrically constrained region, the full mask also includes pixels with no geometric constraint, where multiple different reflections are equally plausible. Because every method --- including ours --- must hallucinate content for these unconstrained pixels, they add cross-seed variance regardless of method, narrowing our relative advantage.

\section{More Qualitative Results}
\label{sec:supp_more_qualitative}

We present additional qualitative comparisons on real captured images without access to ground-truth geometry, complementing \cref{fig:qualitative_results}. All methods are evaluated using identical prompts and seeds. As shown in \cref{tab:more_qualitative_results}, generative baselines often produce visually plausible reflections but exhibit geometric inconsistencies such as incorrect orientation, distorted structure, or misaligned reflected objects. In contrast, our method preserves reflection geometry while maintaining visual realism. \Cref{tab:mirrorbench_v2_qualitative} provides additional comparisons on MirrorBench-V2.

\setlength{\tabcolsep}{2pt}
\renewcommand{\arraystretch}{1}

\begin{table*}[!t]
\centering
\caption{\textbf{Additional qualitative comparison on real captured images without ground-truth geometry.}
Columns (left to right): Ground Truth, MirrorFusion 2.0, FLUX.1 Fill, Qwen-Image-Edit, and our method. Our approach preserves geometric alignment across mirror boundaries and produces consistent reflected structure, while competing methods exhibit orientation errors, structural distortions, or inconsistent occlusions. All methods use identical prompts and the same seeds.}
\label{tab:more_qualitative_results}
\begin{tabular}{ccccc}
\textbf{GT} & \textbf{MF2.0} & \textbf{FLUX.1Fill} & \textbf{Qwen} & \textbf{Ours} \\[0.1cm]
\comparisonrow[0.16]{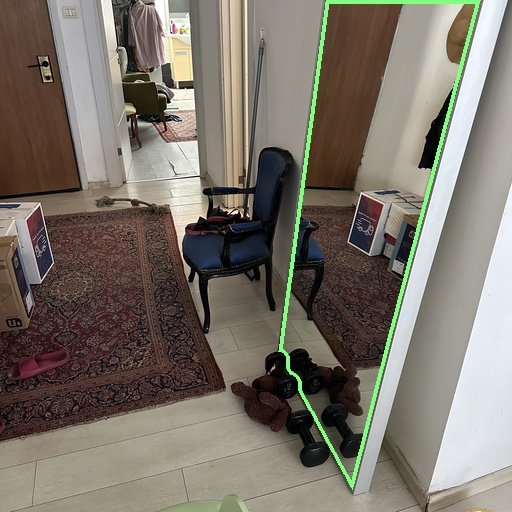}
\comparisonrow[0.16]{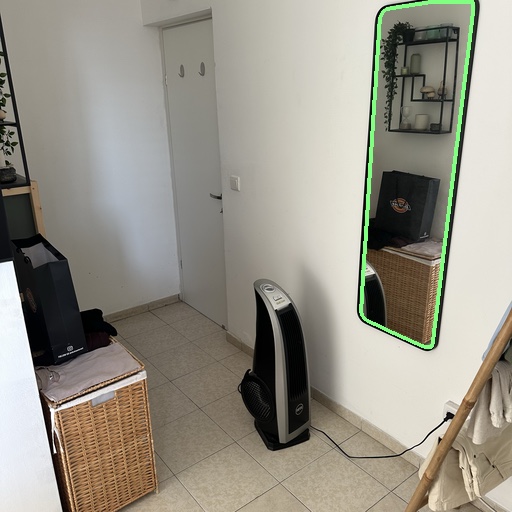}
\comparisonrow[0.16]{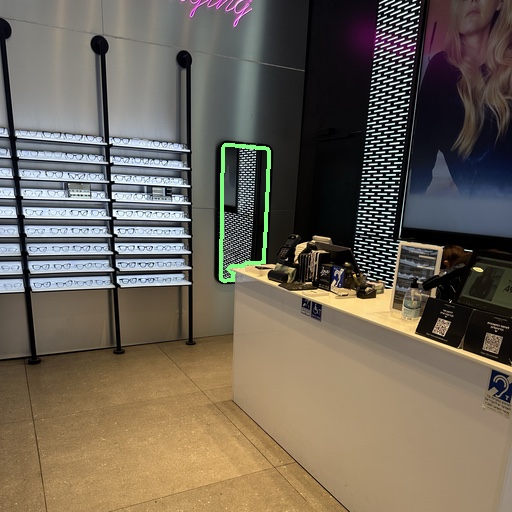}
\comparisonrow[0.16]{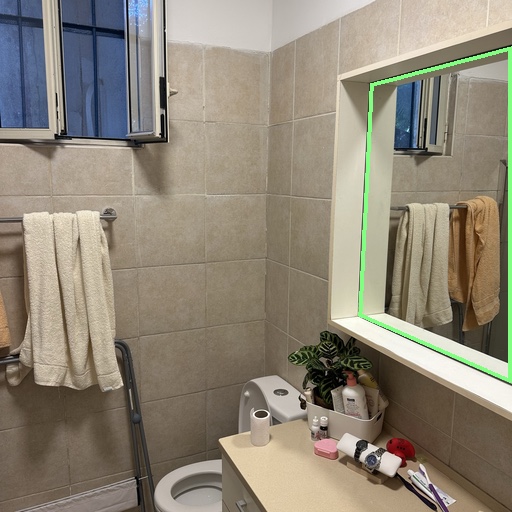}
\comparisonrow[0.16]{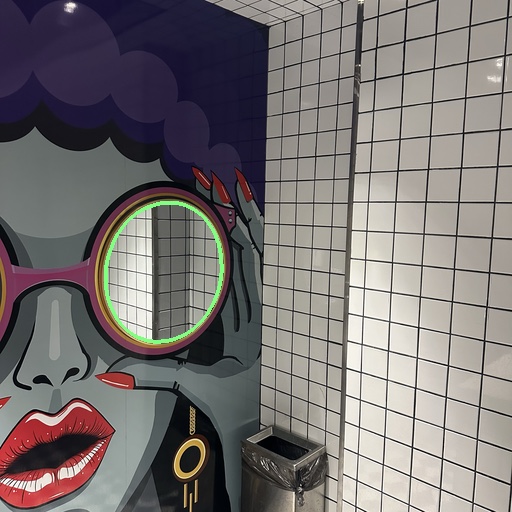}
\comparisonrow[0.16]{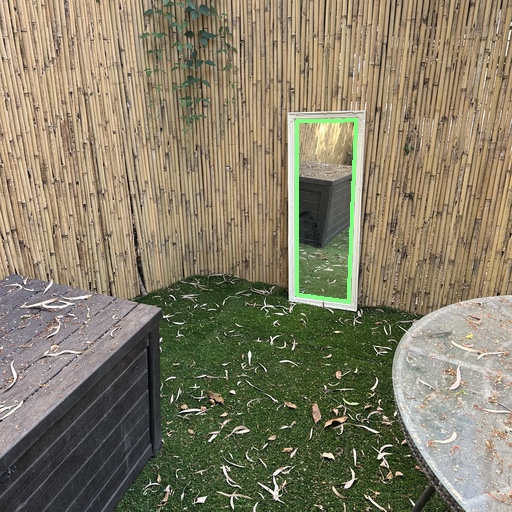}
\end{tabular}
\end{table*}

\begin{table*}[!t]
\ContinuedFloat
\centering
\caption[]{\textbf{Additional qualitative comparison on real captured images without ground-truth geometry (continued).}}
\begin{tabular}{ccccc}
\textbf{GT} & \textbf{MF2.0} & \textbf{FLUX.1Fill} & \textbf{Qwen} & \textbf{Ours} \\[0.1cm]
\comparisonrow[0.16]{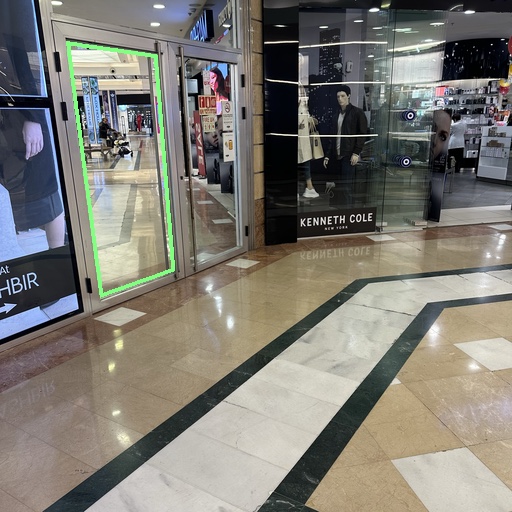}
\comparisonrow[0.16]{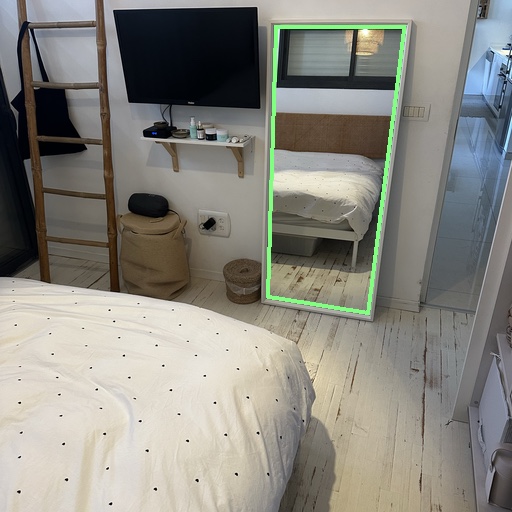}
\end{tabular}
\end{table*}

\setlength{\tabcolsep}{2pt}
\renewcommand{\arraystretch}{1}

\begin{table*}[!t]
\centering
\caption{\textbf{Additional qualitative comparison on MirrorBench-V2.}
Columns (left to right): Ground Truth, MirrorFusion 2.0, FLUX.1 Fill, Qwen-Image-Edit, and our method.}
\label{tab:mirrorbench_v2_qualitative}
\begin{tabular}{ccccc}
\textbf{GT} & \textbf{MF2.0} & \textbf{FLUX.1Fill} & \textbf{Qwen} & \textbf{Ours} \\[0.1cm]
\mirrorbenchcomparisonrow[0.16]{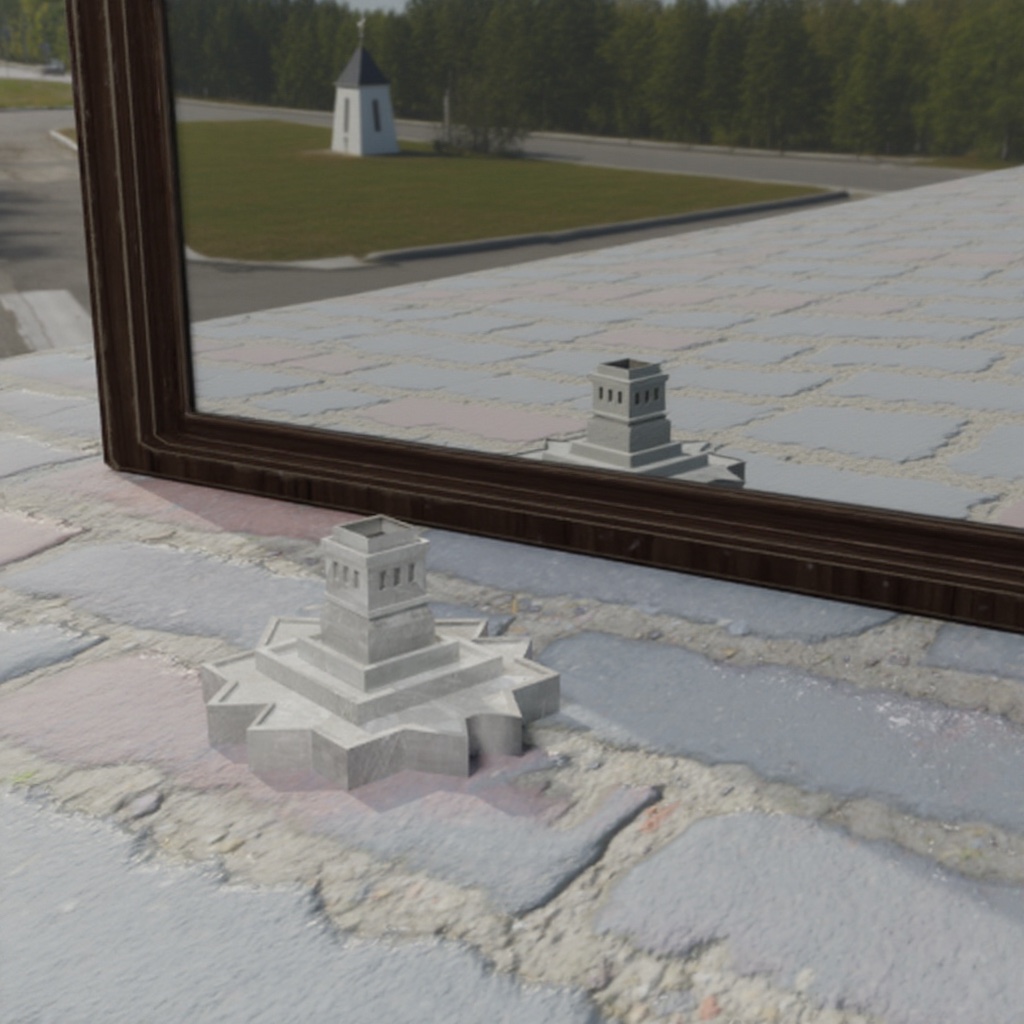}
\mirrorbenchcomparisonrow[0.16]{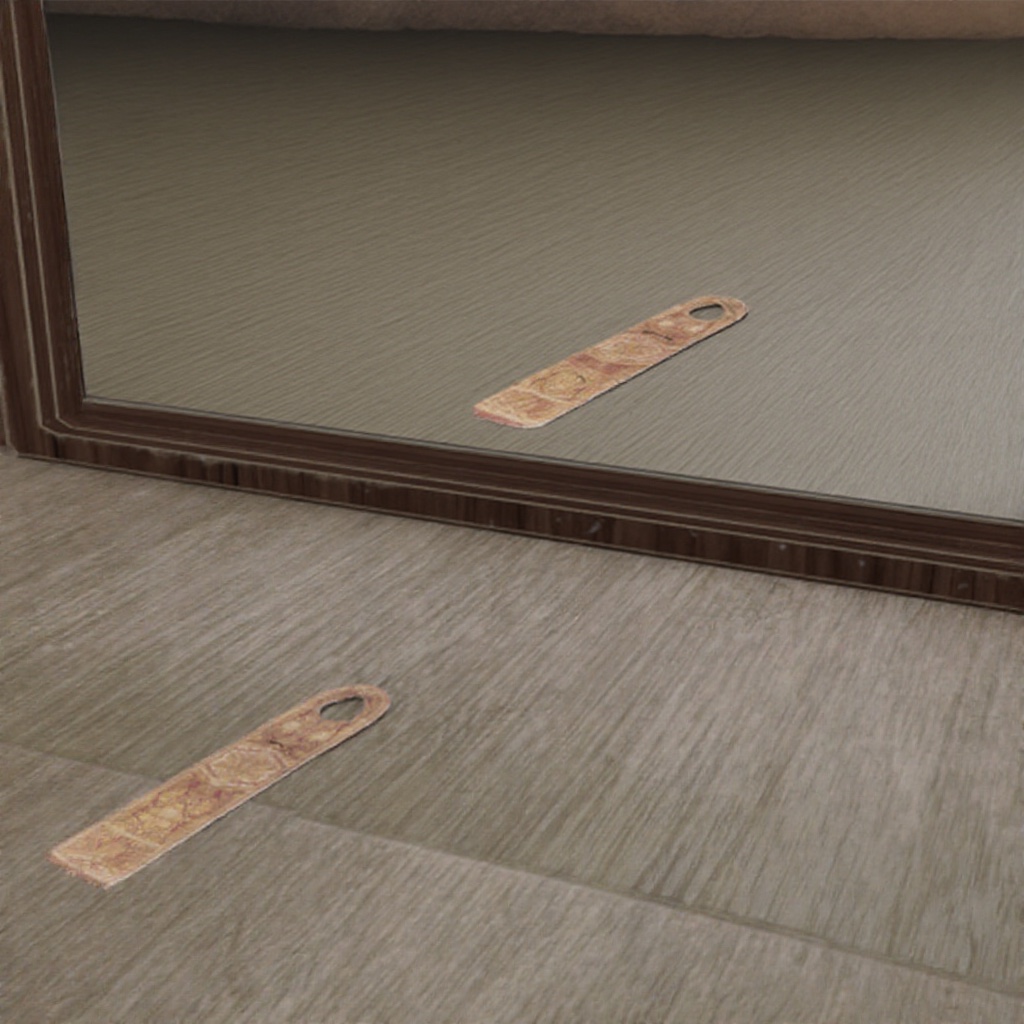}
\mirrorbenchcomparisonrow[0.16]{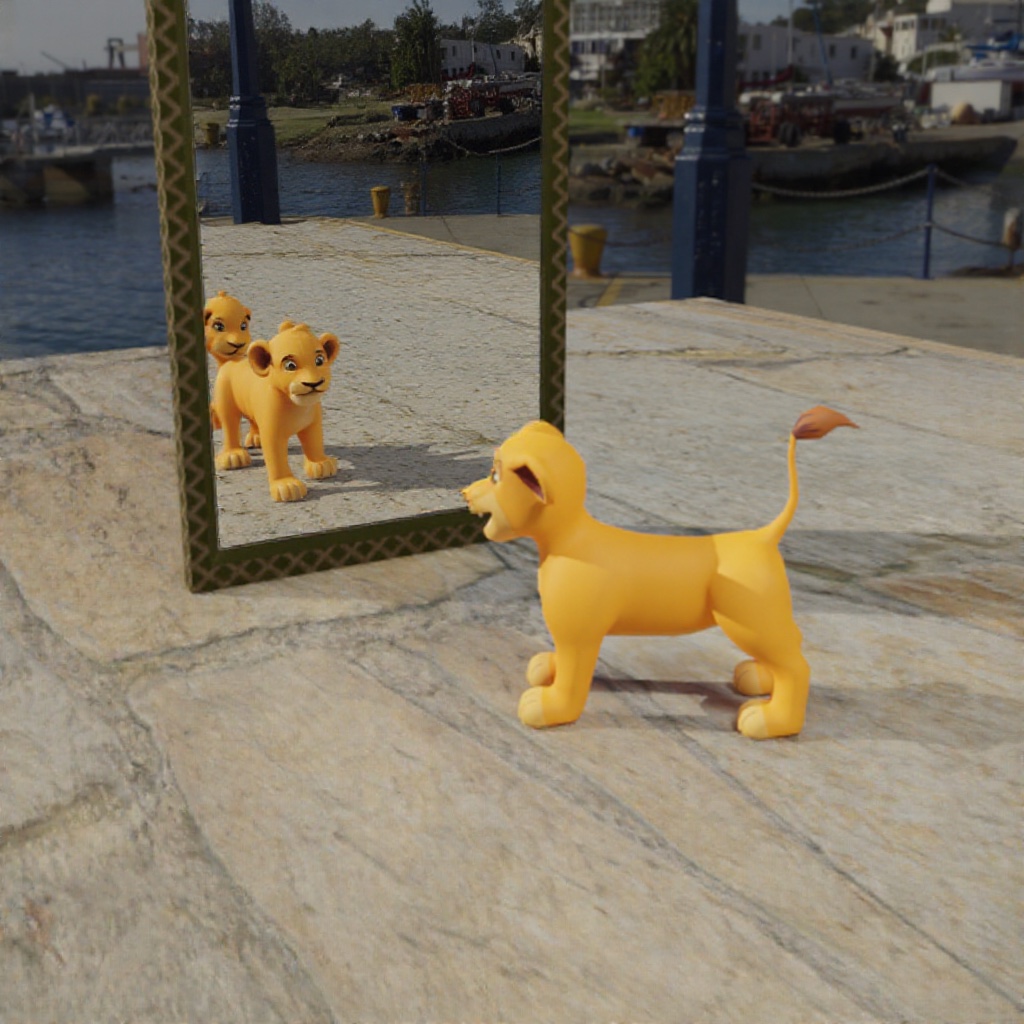}
\mirrorbenchcomparisonrow[0.16]{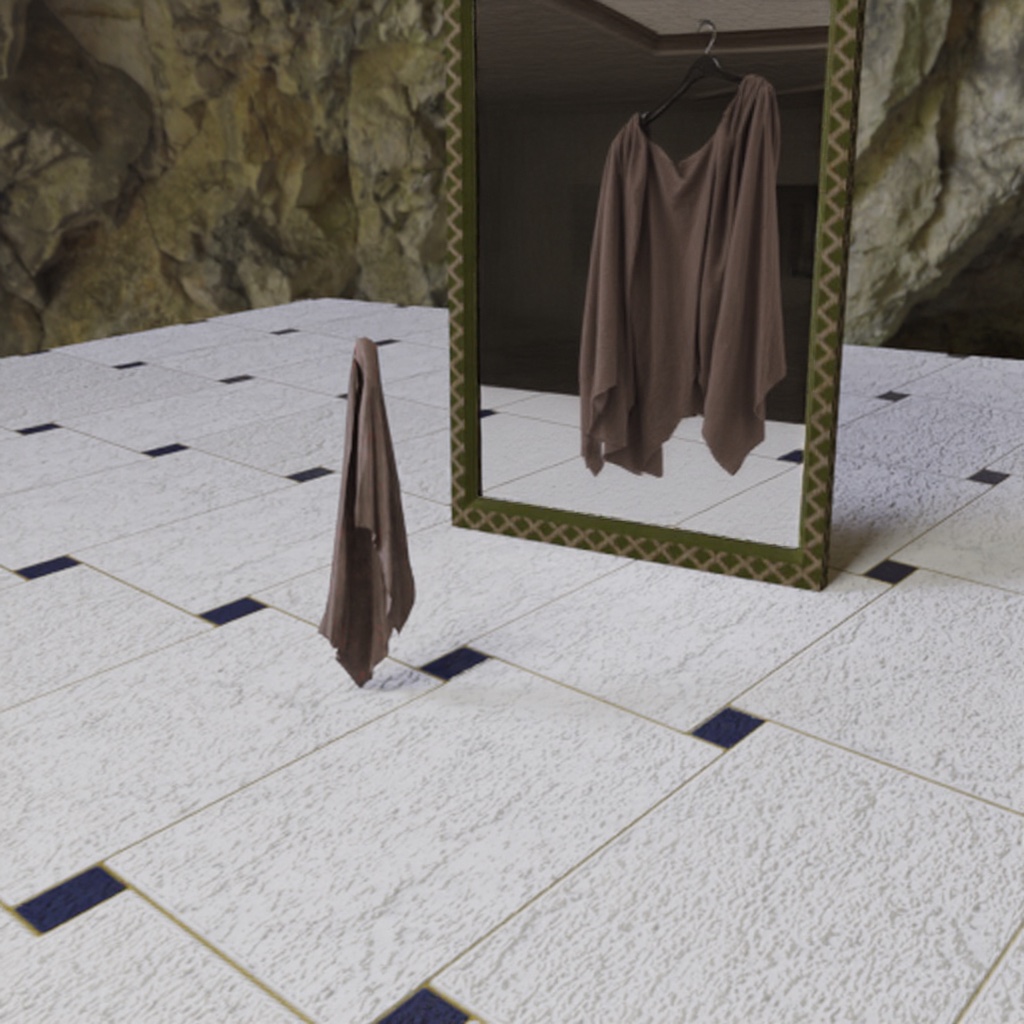}
\end{tabular}
\end{table*}

\begin{table*}[!t]
\ContinuedFloat
\centering
\caption[]{\textbf{Additional qualitative comparison on MirrorBench-V2 (continued).}}
\begin{tabular}{ccccc}
\textbf{GT} & \textbf{MF2.0} & \textbf{FLUX.1Fill} & \textbf{Qwen} & \textbf{Ours} \\[0.1cm]
\mirrorbenchcomparisonrow[0.16]{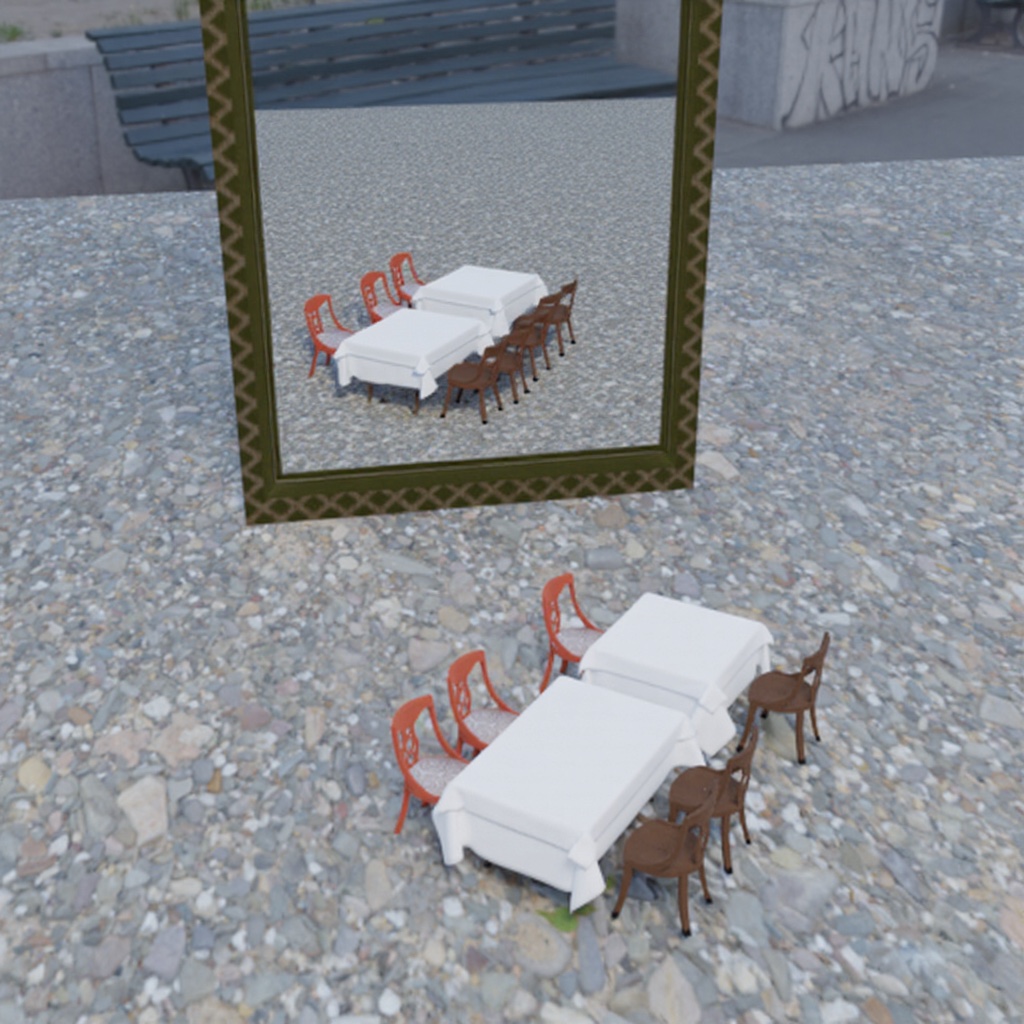}
\mirrorbenchcomparisonrow[0.16]{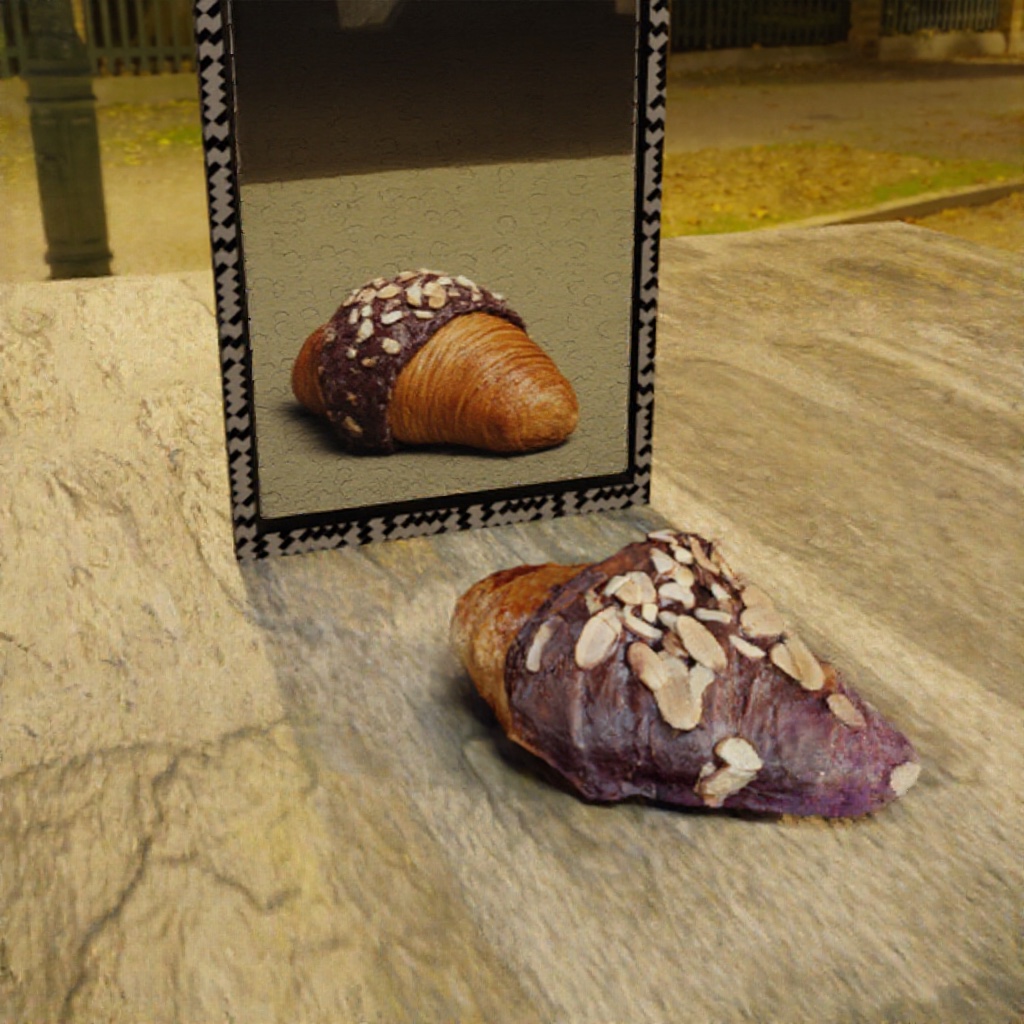}
\mirrorbenchcomparisonrow[0.16]{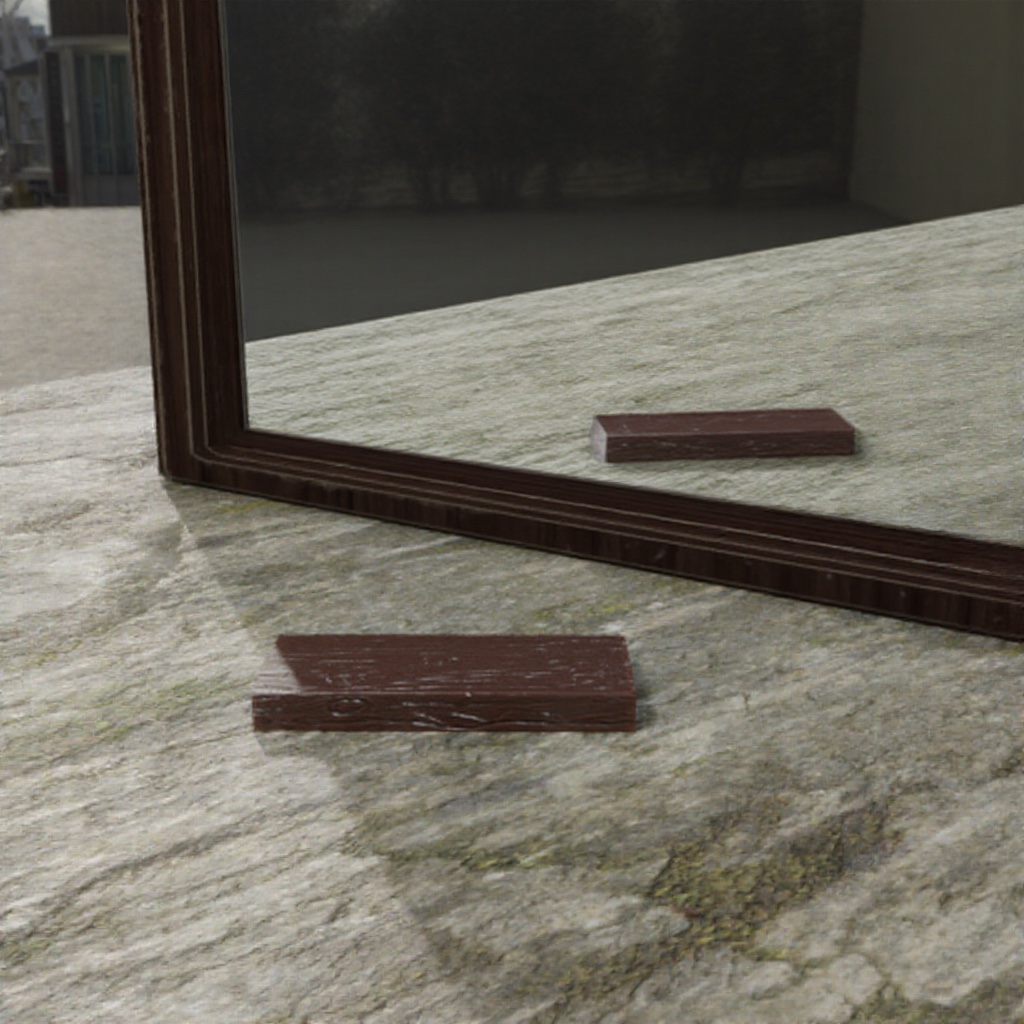}
\end{tabular}
\end{table*}

\section{Robustness to Imperfect Geometry}
\label{sec:supp_robustness}

\begin{figure*}[t]
  \centering
  \begin{subfigure}[b]{0.49\linewidth}
    \includegraphics[width=\linewidth]{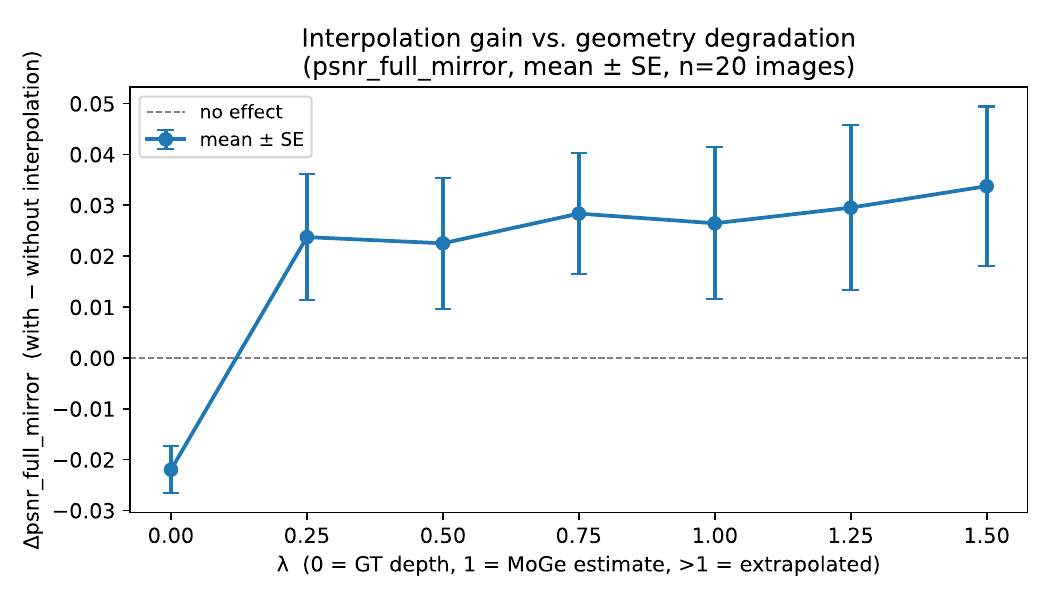}
  \end{subfigure}
  \hfill
  \begin{subfigure}[b]{0.49\linewidth}
    \includegraphics[width=\linewidth]{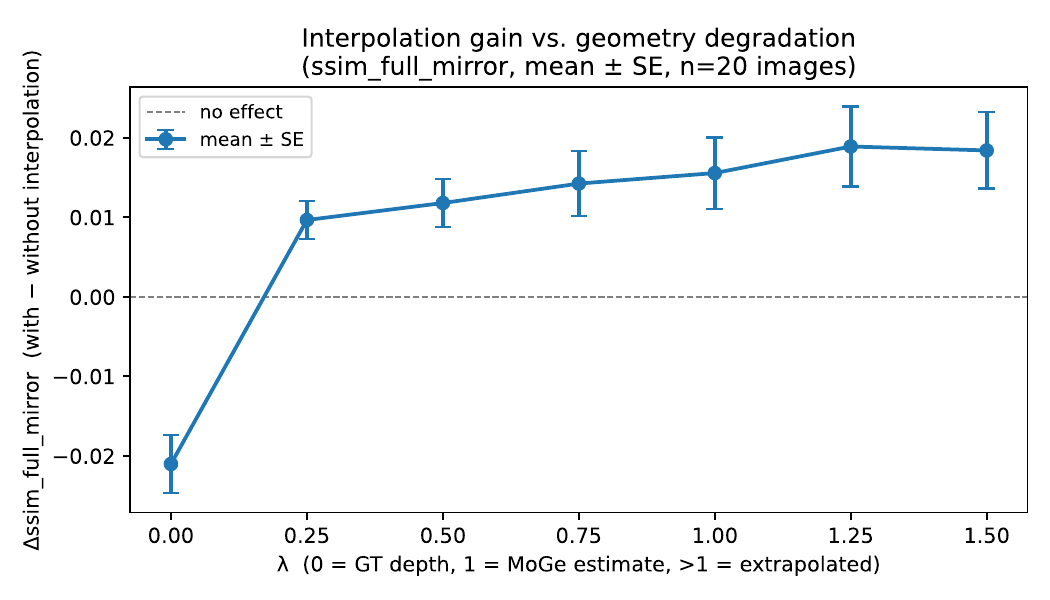}
  \end{subfigure}
  \caption{
    \fromsupp{\textbf{Interpolation gain as geometry degrades.} Gap between our method with and without noise interpolation (mean $\pm$ standard error over 20 MirrorBench-V2 scenes), evaluated on the full mirror mask, as a function of the depth-degradation factor $\lambda$ from \cref{eq:depth_degradation}, for PSNR (\textbf{left}) and SSIM (\textbf{right}). When geometry is exact ($\lambda=0$), interpolation is mildly harmful, slightly departing from an already-correct projection. As soon as geometry deviates from ground truth ($\lambda \geq 0.25$), the gap flips sign, and it continues to grow gradually as $\lambda$ increases further, indicating that interpolation contributes increasingly more as the projected geometry becomes less reliable.}
  }
  \label{fig:depth_degradation_sweep}
\end{figure*}

To examine when noise interpolation helps most, \cref{sec:robustness_main} synthetically degrades the projection input depth (\cref{eq:depth_degradation}) and measures the resulting PSNR/SSIM gap between our method with and without interpolation. \Cref{fig:depth_degradation_sweep} shows this gap as a function of the depth-degradation factor $\lambda$: when geometry is exact, interpolation is mildly harmful, needlessly departing from an already-correct projection, but as projected geometry becomes less reliable it increasingly recovers from occlusion inconsistencies and geometric errors, buying robustness precisely when the projected geometry is least trustworthy.

\section{Additional Robustness: Jitter-Tolerant PSNR}
\label{sec:supp_jitter_psnr}

Standard constrained-region PSNR compares corresponding pixels directly. To measure sensitivity specifically to small local misalignment, we additionally compute \emph{jitter PSNR}: for each constrained ground-truth pixel independently, we search a $\pm3$-pixel neighborhood in the generated image, retain the nearby pixel with the smallest squared error, and average these minimum errors before converting them to PSNR.

\paragraph{Advantages.} First, standard pixel-wise PSNR cannot distinguish incorrect content from correct content shifted by one or two pixels, for example due to minor resampling or interpolation artifacts in the generation pipeline. Jitter PSNR isolates how much of the score is affected by this narrow form of noise rather than treating every one-pixel misalignment as a content error. Second, and critically, our method remains highest under jitter PSNR on both datasets; the ordering is otherwise preserved, except that Qwen-2511 and \MirrorFusion~tie on real images (\cref{tab:jitter_psnr_results}).

\paragraph{Limitation.} Jitter PSNR imposes no coherence constraint across pixels: each pixel independently selects its best local offset, without requiring neighboring pixels to agree on a consistent displacement. It therefore does not verify that content moved together as it would under a physically plausible registration error.
\begin{table*}[t]
\centering
\caption{\textbf{Standard and jitter-tolerant constrained-region PSNR.} Jitter PSNR uses an independent $\pm3$-pixel local search for each evaluated pixel. Our method remains highest on both datasets.}
\label{tab:jitter_psnr_results}
\begin{tabular}{llcc}
\toprule
Dataset & Method & Standard PSNR (dB) $\uparrow$ & Jitter PSNR (dB) $\uparrow$ \\
\midrule
\multirow{5}{*}{\makecell[l]{Real images\\seeds $0,42,512$}}
& \MirrorFusion & 10.12 & 10.74 \\
& Qwen-2511 & 10.16 & 10.74 \\
& \modelName & 13.72 & 15.74 \\
\cdashline{2-4}
& Ours (no interpolation) & \textbf{14.48} & \textbf{16.55} \\
& Ours (with interpolation) & \textbf{14.64} & \textbf{16.67} \\
\midrule
\multirow{5}{*}{\makecell[l]{MirrorBench-V2\\seed $0$}}
& \MirrorFusion & 10.75 & 13.22 \\
& Qwen-2511 & 10.59 & 11.34 \\
& \modelName & 14.87 & 18.96 \\
\cdashline{2-4}
& Ours (estimated geometry) & \textbf{16.35} & \textbf{20.02} \\
& Ours (GT geometry) & \textbf{21.92} & \textbf{25.48} \\
\bottomrule
\end{tabular}
\end{table*}

\FloatBarrier

\end{document}